\pdfoutput=1

\documentclass[11pt]{article}

\usepackage[final]{acl}

\usepackage{times}
\usepackage{latexsym}
\usepackage{enumitem}
\usepackage{hyperref}
\usepackage{dsfont}

\usepackage[T1]{fontenc}

\usepackage[utf8]{inputenc}

\usepackage{microtype}

\usepackage{inconsolata}

\usepackage{graphicx}
\usepackage{adjustbox}

\usepackage[symbol]{footmisc}

\newcommand{\datasetname}{\texttt{XNationQA}}

\title{Investigating Cultural Literacy of Multilingual Large Language Models}
\title{Are Multilingual LLMs Culturally Literate?}
\title{{\em Do You Know About My Nation?}  Investigating Multilingual Language Models’ Cultural Literacy Through Factual Knowledge}

\author{
 Eshaan Tanwar\textsuperscript{1}\quad
 Anwoy Chatterjee\textsuperscript{1}\quad
 Michael Saxon\textsuperscript{2}\thanks{Work completed while affiliated with UC Santa Barbara.}\quad
 Alon Albalak\textsuperscript{3}$^*$\quad\\
 \textbf{William Yang Wang\textsuperscript{4}}\quad
 \textbf{Tanmoy Chakraborty\textsuperscript{1}}\\[0.5em] 
 \textsuperscript{1}Indian Institute of Technology Delhi\quad \textsuperscript{2}University of Washington\\
 \textsuperscript{3}Lila Sciences\quad \textsuperscript{4}University of California, Santa Barbara\\[0.25em]
 \small{\texttt{\{eshaantanwar2000, anwoychatterjee\}@gmail.com}}
}

\usepackage{color, colortbl}
\definecolor{Gray}{gray}{0.9}
\usepackage{multirow,adjustbox}
\usepackage{diagbox}
\newcolumntype{M}[1]{>{\centering\arraybackslash}m{#1}}

\usepackage{amsmath}
\usepackage{amsfonts}
\usepackage{booktabs}

\begin{document}
\maketitle
\renewcommand{\thefootnote}{\arabic{footnote}}
\begin{abstract}
Most multilingual question-answering benchmarks, while covering a diverse pool of languages, do not factor in regional diversity in the information they capture and tend to be Western-centric. This introduces a significant gap in fairly evaluating multilingual models' comprehension of factual information from diverse geographical locations. To address this, we introduce \datasetname~for investigating the \textit{cultural literacy} of multilingual LLMs. \datasetname~encompasses a total of $49,280$ questions on the geography, culture, and history of nine countries, presented in seven languages. We benchmark eight standard multilingual LLMs on \datasetname~and evaluate them using two novel transference metrics. Our analyses uncover a considerable discrepancy in the models' accessibility to culturally specific facts across languages. Notably, we often find that a model demonstrates greater knowledge of cultural information in English than in the dominant language of the respective culture. The models exhibit better performance in Western languages, although this does not necessarily translate to being more literate for Western countries, which is counterintuitive. Furthermore, we observe that models have a very limited ability to transfer knowledge across languages, particularly evident in open-source models\footnote{The source code and \datasetname~are available at \href{https://github.com/EshaanT/XNationQA}{https://github.com/EshaanT/XNationQA}.}.
\end{abstract}

\begin{figure*}[!t]
\begin{center}
\includegraphics[width=0.95\linewidth]{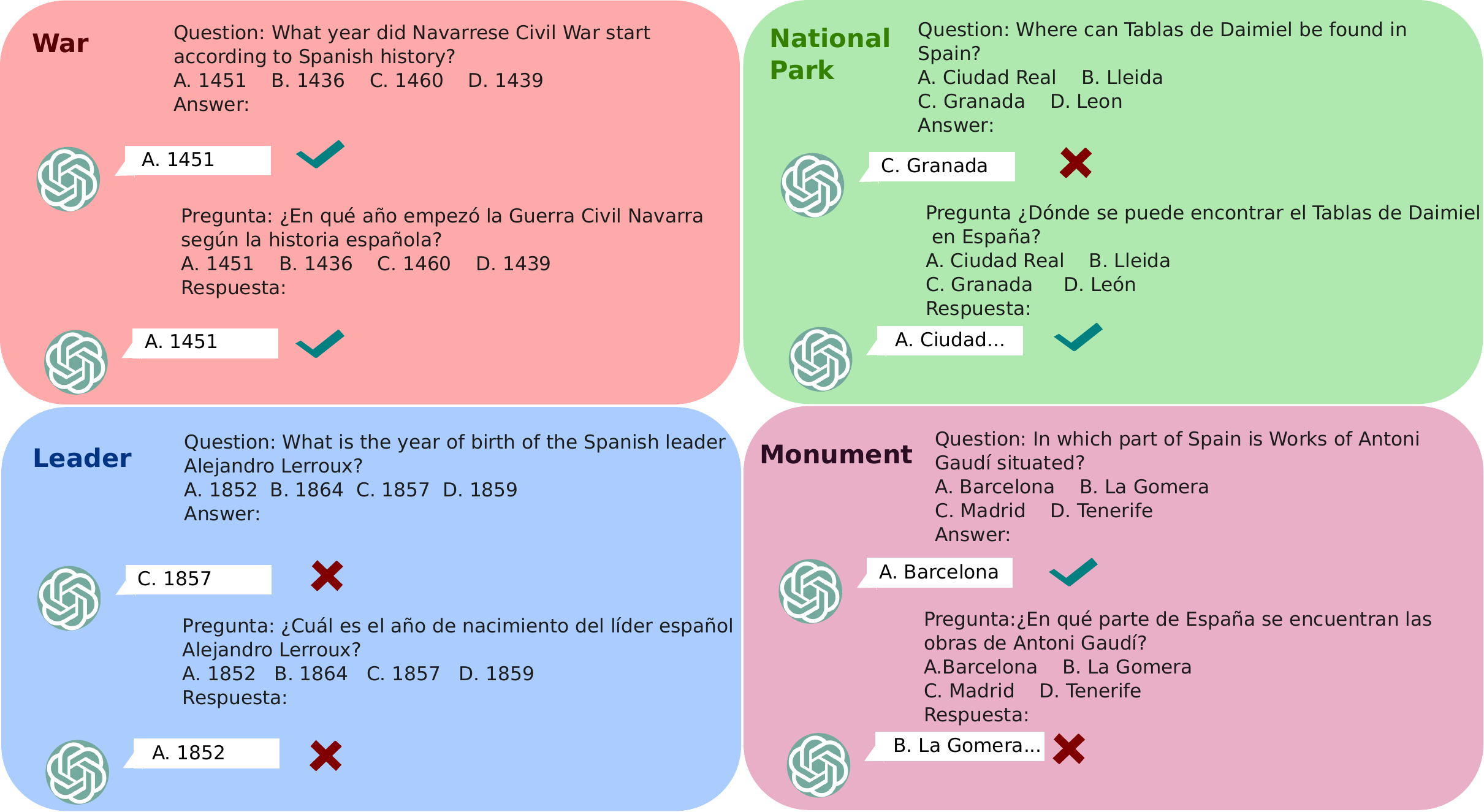}
\caption{An example evaluating an LLM's cultural literacy for Spain, in both English and Spanish. The model answers the \textit{war} question correctly in both English and Spanish but fails on the \textit{leader} question in both. Its performance on the \textit{national park} and \textit{monument} questions is language-dependent, highlighting inconsistencies in its cultural literacy across languages and topics.
}
\label{fig:working-example}
\end{center}
\vspace{-5mm}
\end{figure*}


\section{Introduction}

Multilingual Large Language Models (LLMs) \citep{ustun2024aya,achiam2023gpt} show impressive performance on many languages across varied tasks. However, the best practices to evaluate them remain contested \citep{Ahuja2023MEGAME,Hada2023AreLL,Saha2023BranchSolveMergeIL}, with
many criticising Western-centric (often Anglosphere) evaluations~\citep{Held2023AML}. 
For example, \citet{faisal2021dataset} demonstrated how multiple purportedly multilingual question-answering (QA) datasets disproportionately cover USA-related concepts and entities, with a similar question distribution to monolingual English benchmarks.
This is because multilingual QA benchmarks are often derived from Western-centric English datasets~\citep{longpre2021mkqa,kassner-etal-2021-multilingual,dumitrescu2021liro}. 
They, thus, fail to consider the cultural contexts where these languages are spoken \citep{Naous2023HavingBA}. 
\citet{Blevins2022AnalyzingTM} proposed guidance techniques to create more inclusive benchmarks. To the best of our knowledge, there is no easy-to-evaluate large-scale parallel multilingual QA dataset that explicitly balances its distribution across a set of diverse cultures, capturing factual knowledge. 
LLMs are clearly able to access factual knowledge in multiple languages \citep{jiang-etal-2020-x,kassner-etal-2021-multilingual}, but the relationship between language and information locality is poorly understood.
To this end, and motivated by the education research literature, we produce a parallel knowledge test, \textit{consisting of regionally differing factual information}, balanced across a set of related factual queries and languages.

In education theory, \textit{cultural literacy} refers to 
the shared corpus of translinguistic cultural knowledge within a community, polity, or society on which efficient communication is built \citep{hirsch1983cultural}. 
By \textit{translinguistic},~\citet{hirsch1983cultural} meant that the information in the corpus is language-agnostic in its meaning. An interesting study conducted by~\citet{steffensen1979cross} in this regard found that American English speakers performed worse than Indian English speakers on a reading comprehension test of a story describing an Indian wedding, and vice versa on a story of an American one at equivalent reading levels because the context was more familiar to the speakers of the corresponding culture. Multilingual LLMs strive to create technology that is inclusive of a diverse set of linguistic groups~\citep{Blasi2021SystematicII}. Therefore, they need to be culturally literate across the languages they cover to consistently answer translinguistic regional questions on specific topics concerning different nations.

Using cultural literacy as a framing device, we aim to investigate multilingual and multicultural knowledge in multilingual LLMs by answering the following research questions:

\noindent\textit{\textbf{RQ1:}} \textit{What disparities exist in the cultural literacy of various LLMs about different countries?}

\noindent\textit{\textbf{RQ2:}} \textit{How transferable is the cultural literacy of LLMs across different languages?}


\noindent\textit{\textbf{RQ3:}} \textit{How consistent are LLMs in recalling facts about nations across different languages?}

\noindent\textit{\textbf{RQ4:}} \textit{Does a systematic relationship between country and language exist in LLMs' cultural literacy?}


To facilitate this investigation, we create \datasetname, a test set designed to assess cultural literacy through QA pairs that evaluate curricular knowledge for nine countries (Japan, India, China, Germany, Spain, Russia, Mexico, the USA, and the UK), fully translated into seven languages (Japanese, Hindi, Chinese, German, Spanish, Russian, and English). The dataset covers a set of four axes, namely: questions about wars, leaders, monuments, and national parks. Given culture's nuanced and context-dependent nature, which can vary across communities, languages, and sociopolitical environments, it is inherently challenging to define and quantify. To operationalize cultural literacy in our study, we draw on prior work in the monolingual (English-centric) setting that measures geographic and cultural erosion in language models via factual recall~\cite{schwobel-etal-2023-geographical,zhou-etal-2022-richer}. Following this approach, we assess a model’s cultural literacy about a nation through its performance on questions concerning historically and geographically grounded facts specific to that nation. We categorise these facts into four domains: (1) \textit{wars}, or the year of armed conflicts undertaken by the country,
(2)  \textit{national parks}, or the site of nationally designated protected wilderness areas in each country,
(3) \textit{leaders}, or the date of birth of heads of state of the country,
and (4) \textit{monuments}, or the locations of UNESCO world heritage sites in the country. This framing allows us to systematically evaluate a model’s knowledge of culturally embedded facts, serving as a proxy for its cultural literacy. 
Figure \ref{fig:working-example} shows a working example of our task. We prompt the model to answer a question each on monuments, wars, leaders, and national parks. The example highlights the model's inconsistent cultural literacy across languages. For instance, the model answers the war question correctly in both English and Spanish but fails the leader question in both. Meanwhile, its performance on the national park and monument questions is language-dependent, answering correctly in one language but not the other.

We extensively evaluate eight commonly-used multilingual LLMs on \datasetname\ and find that:
\begin{itemize}[nosep, wide, labelwidth=!, labelindent=0pt]
    \item \textit{LLMs show a disparity in cultural literacy across languages}, with models exhibiting varying knowledge. Models tend to perform the best in English, followed by other Western languages (German, Spanish, and Russian).
    \item \textit{LLMs have poor transference of facts across languages}, with our novel coverage metrics showing that open-source models struggle to answer faithfully in all languages. When considering only Western languages, models tend to be more faithful, indicating a knowledge coverage disparity between Western and non-Western languages.
    \item \textit{LLMs exhibit varying performance across nations}, with different models exhibiting different levels of literacy about different nations.  Surprisingly, a nation's native language is not the best-performing language even for questions on that nation, with Western languages performing the best.
\end{itemize}

We also observe that although the models generally perform better in Western languages, this does not translate to higher accuracy for Western countries; in fact, models sometimes exhibit greater cultural literacy for countries like India, China, and Japan than for Western nations such as Germany or Spain.

\section{Related Work and Motivation}

\paragraph{Multilingual Language Models.} Since the advent of pre-trained transformer-based language models~\citep{devlin2018bert}, there has been a constant effort to develop multilingual LLMs that can understand and reason in multiple languages. These variants are trained on unsupervised training objectives using large multilingual corpora such as Oscar~\citep{2022arXiv220106642A}, mC4~\citep{xue-etal-2021-mt5}, and CulturaX~\citep{nguyen2023culturax}, which are not parallel across languages. Therefore, the multilingual generalization ability of these models is a by-product of their ability to project different languages into a common representation space~\citep{Artetxe2019OnTC,Blevins2022AnalyzingTM}. This representation ability directly depends on the datasets they are trained on~\citep{deshpande-etal-2022-bert}. However, studies have shown them to be biased towards Western concepts due to their training data mixture~\citep{Naous2023HavingBA,Cao2023AssessingCA}.

\paragraph{Multilingual Benchmarks.} Over the years, multiple multilingual benchmarks spanning various NLP tasks~\citep{dac2023okapi} have been created to evaluate the multilingual abilities of language models. However, these benchmarks are commonly derived from monolingual English benchmarks~\citep{clark2020tydi,dumitrescu2021liro}, and, hence, tend to be biased in their coverage. Our initial analysis using multilingual QA datasets reveals these problems. For example, MKQA~\citep{longpre2021mkqa}, a popular closed-book QA dataset, is generated by translating $10,000$ samples from Google's Natural Questions dataset~\citep{kwiatkowski2019natural} without any cultural or regional considerations. We find the resultant dataset to contain questions about popular Hollywood shows like `Modern Family' and `Rick and Morty', while it does not cover Bollywood or other regional shows. Hence, these datasets fail to highlight if the model is culturally literate across languages and regions. In contrast, \datasetname~is large ($49,280$ questions) and covers multiple specific topics in parallel sets for nine nations.

\paragraph{Factual and Cultural Knowledge in Multilingual Models.}
Several recent studies have begun to probe cultural knowledge in multilingual settings~\cite{fung2024massively,shi2024culturebank}. \citet{keleg2023dlama} explores multilingual factual knowledge by mining Wikidata triples in four languages; however, its reliance on language-linked labels restricts coverage to only entities available in all languages -- a limitation that overlooks culturally important items, like the Mahatma Gandhi Marine National Park, and thus lacks uniform representation. \datasetname~addresses such limitations by employing translation toolkits to construct a parallel corpus across seven languages, ensuring equitable coverage of historically and culturally significant content. While other works like BLEnD~\cite{myung2024blend} and CaLMQA~\cite{arora2024calmqa} also emphasise cultural specificity, they differ from our work in scope and evaluation paradigm. BLEnD focuses on subjective, everyday knowledge sourced from native speakers, while CaLMQA uses a generative format for long-form questions that necessitates high-budget human evaluation. In contrast, \datasetname~targets objective, historically anchored factual knowledge that is widely documented (e.g., wars, national parks, monuments, leaders) through a multiple-choice setup that enables scalable and accessible benchmarking for future studies.

\section{\datasetname ~Dataset}

\datasetname~is a parallel multilingual dataset encompassing factual knowledge related to nine countries and seven languages. \datasetname~ is designed to evaluate the cultural literacy of LLMs across languages by testing their knowledge of nation-specific information. Each instance within \datasetname~ includes an objective-type question focusing on a specific domain, with one correct and three incorrect options. In total, the dataset contains $49,280$ questions, spanning $1,760$ factual entities. Figure~\ref{fig:pie} shows the distribution of the entities across nine countries we covered.


\paragraph{Mining Entities.} 
We began by first mining factual entities specific to a country. We used English Wikipedia pages to obtain lists for four specific domains: (i) leaders of a country, (ii) national parks in a country, (iii) UNESCO sites in the nation, and (iv) wars the nation has participated in. This method was applied to extract entities for nine countries: Japan, India, China, Germany, Spain, Russia, Mexico, the USA, and the UK (see Appendix~\ref{sec:dataset_dis} for further details).

After extracting the entity list,  we used Wikidata to mine additional information. This included information about the start year of a war, the birth year of a leader, and the administrative location of a national park and the UNESCO site. Information about the entities without an appropriate Wikidata entry was filled out manually. This process produced the final entity-answer pairs used to construct the questions.

\paragraph{Constructing Question Templates and Options.} After extracting a nation's entity-answer pair, we manually created four prompt templates for each domain and country in English. Each domain's template was designed to ask a specific question, either about the year or about the location (refer to Table \ref{tab:examples} of Appendix~\ref{sec:prompt_tempt} for details on the prompt templates for each domain). 

To generate the desired objective-type questions, we generated three incorrect answers by sampling three random administrative areas of a country where the UNESCO site and national park are not located. Additionally, for year-type questions, we generated incorrect answers by randomly adding or subtracting a number between $5$ and $10$ from the correct year. These incorrect answers, along with the correct answer, are paired with the manually created question templates associated with an entity to form multiple-choice questions. Hence, for each entity, we generate four prompts. This approach is designed to account for any prompt sensitivity or bias in the LLMs.

\paragraph{Expansion to Other Languages.} We use translation toolkits to create the parallel multilingual corpus. This is done because not all entities in Wikidata have labels in multiple languages. Hence, to create \datasetname, all the generated templates and entity-option pairs were translated using Google Translate and GPT-4, respectively, into Hindi, Spanish, Chinese, Japanese, Russian, and German. We filled out the entity and the options in the templates in their respective languages to generate the relevant objective question. This resulted in a geographically diverse, parallel multilingual factual knowledge corpus, \datasetname.

\begin{table}[!t]
\centering
\adjustbox{max width=1\linewidth}{
\begin{tabular}{l|cccccc|c}
\hline
\textbf{Score} & \textbf{DE} & \textbf{ES} & \textbf{HI} & \textbf{RU} & \textbf{JA} & \textbf{ZH} & \textbf{AVG} \\ \hline \hline
Cos-sim&$0.94$&$0.94$&$0.87$&$0.90$&$0.87$&$0.87$&$0.89$\\
BLEU&$50.80$&$67.45$&$68.48$&$43.80$&$52.30$&$45.64$&$54.74$\\
H-Eval&$4.85$&$4.75$&$4.57$&$4.25$&$4.50$&$4.41$&$4.55$\\
\hline
\end{tabular}
}
\caption{Validation of dataset quality using human evaluation (H-Eval), BLEU score, and cosine-similarity (Cos-sim). The queries seem to be semantically aligned (see Table \ref{tab:langs_iso} of Appendix~\ref{appendix:iso} for the languages corresponding to the ISO codes).}
\label{tab:valid_dataset}
\vspace{-5mm}
\end{table}

\paragraph{Validation of Translation.} 
To validate the quality of \datasetname, we employ back-translation~\citep{Miyabe2015EvaluationOT} and semantic similarity (c.f. Table~\ref{tab:valid_dataset}). We also conduct human evaluation on a subset of queries to further ensure that the translation quality is preserved across languages.
\begin{enumerate}
    \item[(i)] \textbf{Semantic Similarity: }To ensure that queries in different languages have the same semantic meaning as their English counterparts, we compute the cosine similarity between the English queries and their translations. We use multilingual sentence transformers to extract embeddings, which are then used to compute cosine similarity. Our dataset has an average similarity score of $0.89$, suggesting cross-lingual consistency and meaning preservation.  
    
    \item[(ii)] \textbf{Back-Translation: }In this study, we randomly selected $1,000$ queries from \datasetname{} for each of the translated languages, i.e., $6,000$ queries in total, across the various topics in our dataset. These queries were back-translated into English using Google Translate and then compared to their original English versions. We observed an average BLEU score of $54.74$ which indicates that the translations preserve the concepts of the original queries.

    \item[(iii)] \textbf{Human Evaluation: }To assert the grammatical correctness and overall clarity of our dataset, $1,000$ queries (same as in back-translation) from each language in~\datasetname, i.e., $6,000$ queries in total, were evaluated by language experts. The experts assigned a score of $1$ to $5$ based on grammar, fluency, and coherence to assess the quality of the dataset. We found that all samples at least got a score of $4$ with an average score of $4.55$, indicating the high quality of our dataset (see Appendix~\ref{sec:HE} for further details on the human evaluation procedure).
\end{enumerate}

\section{Problem Definition and Experimental Setup}

Our dataset $\mathcal{D}$ spans $L$ languages and covers a set of nation-specific factual entities $\mathcal{E}$. For each entity $e\in \mathcal{E}$ in our dataset, we have a set of four manually created questions $q^l$ and options $o^l$ where language $l \in L$. The task is to generate the correct answer $a^l$ from the options. To do so, we formalize our prompting setup as generating the output $\hat{y}^l$ conditioned upon the question and option, i.e., $\hat{y}^l=\mathrm{argmax}\, P(y|q^l \oplus o^l)$.
We then match the generated output $\hat{y}^l$ with the correct answer $a^l$ across all languages $l$ and entities $e$, to check for transferability of cultural literacy.

We experiment with eight commonly-used instruction-tuned multilingual LLMs, specifically the 7 and 13-billion versions of LLaMA-2-Chat~\citep{touvron2023llama}, 8-billion Meta-LLaMA-3-Instruct~\citep{dubey2024llama}, 7-billion Bloomz~\citep{yong-etal-2023-bloom}, 7-billion Mistral-Instruct~\citep{jiang2023mistral}, 7-billion Mixtral~\citep{jiang2024mixtral}, 13-billion Aya~\citep{ustun2024aya} and GPT-4~\citep{achiam2023gpt}. For the GPT-4 model, due to budgetary constraints, we sample one question for each entity in every language. In total we evaluated GPT-4 on $12,320$ questions. These models have different mixtures of languages in their training corpus (refer to Appendix~\ref{sec:model_language_apendix} for more information). Further, we also extend our analyses to Meta-LLaMA-3.1-8B-Instruct~\citep{meta2024llama3.1} and Qwen3 (8B and 14B)~\citep{qwen3} models --- the findings of which are detailed in Appendix~\ref{sec:eval_qwen_llama3.1_appendix}.

\begin{table*}[!t]
\centering
\resizebox{\textwidth}{!}{ 
\begin{tabular}{lccccccc|ccc}
\hline
 \backslashbox{Model}{Language}& $EN$& $DE$& $ES$& $HI$& $RU$& $JA$ &$ZH$ & $AVG$& $AVG_{W}$&$AVG_{NW}$\\ \hline
\multicolumn{11}{c}{\cellcolor[HTML]{EEEEEE}Monuments}
 \\ \hline
Bloomz-7B1&$57.30$&$45.64$&$51.51$&$21.00$&$13.08$&$17.88$&$44.04$&$35.78$&$41.88$&$27.64$\\
LLaMA-2-7B-Chat&$82.03$&$75.98$&$76.25$&$3.83$&$36.30$&$37.72$&$34.43$&$49.50$&$67.64$&$25.31$\\
Mistral-7B-Instruct&$88.26$&$80.07$&$81.41$&$29.27$&$65.30$&$26.25$&$33.54$&$57.73$&$78.76$&$29.68$\\
Meta-LLaMA-3-8B-Instruct&$59.16$&$88.52$&$87.90$&$53.91$&$57.38$&$55.43$&$63.08$&$66.48$&$73.24$&$57.46$\\
LLaMA-2-13B-Chat&$85.41$&$77.76$&$80.16$&$15.39$&$54.80$&$37.19$&$31.49$&$54.60$&$74.53$&$28.02$\\
Aya&$64.15$&$62.81$&$59.52$&$35.68$&$30.34$&$34.34$&$32.83$&$45.67$&$54.20$&$34.29$\\
GPT-4&$\textbf{96.09}$&$\textbf{95.73}$&$\textbf{96.44}$&$\textbf{87.90}$&$\textbf{92.02}$&$\textbf{92.17}$&$\textbf{80.78}$&$\textbf{89.88}$&$\textbf{92.08}$&$\textbf{86.94}$\\
Mixtral-8x7B&$93.33$&$85.50$&$88.52$&$20.46$&$82.12$&$37.19$&$37.19$&$63.47$&$87.37$&$31.60$\\
\hline
\multicolumn{11}{c}{\cellcolor[HTML]{EEEEEE}Leaders} \\ \hline
Bloomz-7B1&$22.25$&$26.83$&$20.08$&$21.67$&$24.00$&$26.58$&$26.00$&$23.92$&$23.29$&$24.76$\\
LLaMA-2-7B-Chat&$56.42$&$46.58$&$27.42$&$8.50$&$23.00$&$28.33$&$28.00$&$31.18$&$38.35$&$21.62$\\
Mistral-7B-Instruct&$32.83$&$34.92$&$33.67$&$22.92$&$32.83$&$19.08$&$28.42$&$29.24$&$33.56$&$23.47$\\
Meta-LLaMA-3-8B-Instruct&$80.25$&$82.92$&$80.42$&$57.58$&$63.08$&$35.50$&$28.67$&$61.20$&$76.67$&$40.57$\\
LLaMA-2-13B-Chat&$52.83$&$30.42$&$30.25$&$18.25$&$26.42$&$25.42$&$27.50$&$30.15$&$34.98$&$23.70$\\
Aya&$23.33$&$22.42$&$21.08$&$22.08$&$23.42$&$18.75$&$19.92$&$21.57$&$22.56$&$20.25$\\
GPT-4&$62.33$&$62.33$&$63.33$&$\textbf{47.33}$&$55.67$&$\textbf{48.00}$&$\textbf{48.33}$&$55.33$&$60.92$&$\textbf{47.87}$\\
Mixtral-8x7B&$\textbf{89.17}$&$\textbf{85.25}$&$\textbf{85.92}$&$9.33$&$\textbf{79.33}$&$32.75$&$28.08$&$\textbf{58.55}$&$\textbf{84.92}$&$23.39$\\ 
\hline
\multicolumn{11}{c}{\cellcolor[HTML]{EEEEEE}Wars} \\ \hline
Bloomz-7B1&$32.93$&$25.11$&$29.04$&$33.96$&$28.12$&$28.79$&$39.24$&$31.02$&$28.80$&$33.98$\\
LLaMA-2-7B-Chat&$56.27$&$46.07$&$42.25$&$24.93$&$34.49$&$39.27$&$35.94$&$39.89$&$44.77$&$33.38$\\
Mistral-7B-Instruct&$48.19$&$45.11$&$46.46$&$25.78$&$44.51$&$30.74$&$35.87$&$39.52$&$46.07$&$30.78$\\
Meta-LLaMA-3-8B-Instruct&$71.74$&$64.09$&$58.64$&$51.17$&$53.54$&$50.74$&$49.43$&$57.05$&$62.00$&$50.78$\\
LLaMA-2-13B-Chat&$47.59$&$40.37$&$34.63$&$20.75$&$37.96$&$37.25$&$35.38$&$36.28$&$40.14$&$31.13$\\
Aya&$30.59$&$27.05$&$27.58$&$26.17$&$26.24$&$26.81$&$24.68$&$27.02$&$27.87$&$25.88$\\
GPT-4&$74.08$&$69.41$&$68.98$&$\textbf{57.79}$&$\textbf{67.85}$&$\textbf{60.34}$&$\textbf{59.07}$&$\textbf{65.36}$&$70.08$&$\textbf{59.06}$\\
Mixtral-8x7B&$\textbf{76.77}$&$\textbf{73.65}$&$\textbf{70.64}$&$26.27$&$66.29$&$49.04$&$47.52$&$58.60$&$\textbf{71.84}$&$40.94$\\ 
\hline
\multicolumn{11}{c}{\cellcolor[HTML]{EEEEEE}National Parks} \\ \hline
Bloomz-7B1&$48.04$&$40.06$&$46.30$&$25.95$&$15.49$&$20.08$&$25.11$&$31.58$&$37.47$&$23.72$\\
LLaMA-2-7B-Chat&$74.63$&$64.80$&$66.12$&$11.42$&$33.14$&$30.66$&$27.85$&$44.09$&$59.67$&$23.31$\\
Mistral-7B-Instruct&$77.33$&$72.94$&$67.39$&$19.50$&$51.11$&$13.90$&$25.69$&$46.84$&$67.19$&$19.70$\\
Meta-LLaMA-3-8B-Instruct&$64.22$&$79.60$&$80.02$&$51.32$&$56.77$&$39.11$&$40.96$&$58.86$&$70.15$&$43.80$\\
LLaMA-2-13B-Chat&$75.53$&$68.97$&$66.70$&$10.15$&$42.49$&$31.9$2&$26.27$&$46.01$&$63.42$&$22.79$\\
Aya&$50.32$&$45.30$&$44.71$&$36.21$&$33.30$&$32.66$&$24.42$&$38.13$&$43.41$&$31.09$\\
GPT-4&$\textbf{97.25}$&$\textbf{94.93}$&$\textbf{94.29}$&$\textbf{83.09}$&$\textbf{89.64}$&$\textbf{74.63}$&$\textbf{60.04}$&$\textbf{84.84}$&$\textbf{94.03}$&$\textbf{72.58}$\\
Mixtral-8x7B&$91.01$&$82.98$&$84.25$&$19.93$&$73.26$&$31.08$&$24.84$&$58.19$&$82.88$&$25.27$\\ 
\hline
\end{tabular}
}
\caption{Model accuracy on \datasetname~across seven languages, averaged over all nine countries. The $AVG_W$ and $AVG_{NW}$ columns show mean accuracy for Western (EN, DE, ES, RU) and non-Western (HI, JA, ZH) languages respectively. While GPT-4 demonstrates the strongest overall cultural literacy, other competitive models like Mixtral-8x7B show a significant drop in performance for non-Western languages (see Table \ref{tab:langs_iso} for the ISO codes).
}
\label{tab:main_results}
\vspace{-5mm}
\end{table*}

\section{Results and Analyses}
\label{sec:acc_results}

\subsection{Cultural Literacy Across Languages}Table~\ref{tab:main_results} presents the accuracy of eight multilingual LLMs on \datasetname~across languages, averaged over countries. The evaluation reveals notable disparities in the cultural literacy of the LLMs. The models, ranked by average accuracy, are GPT-4 ($72\%$), Mixtral-8x7B ($60\%$), LLaMA-3-8B-Instruct ($59\%$), Mistral-7B-Instruct ($42.6\%$), LLaMA-2-7B-Chat ($41\%$), LLaMA-2-13B-Chat ($40\%$), Aya ($32\%$), and Bloomz ($30\%$). 

A key finding is that the performance of the models varies significantly with both the topic and the language of the query. Models are generally more proficient at recalling locations for \textit{monuments} and \textit{national parks} than they are at recalling specific years for \textit{wars} or \textit{leaders} (see Appendix~\ref{sec:model_wise_analysis} for a detailed breakdown). Across all topics, English is typically the best-performing language. 

Surprisingly, models specifically trained for multilingual alignment, such as Bloomz ($46$ languages) and Aya ($101$ languages), underperform compared to the models from LLaMA and Mistral families of comparable size, which have a much smaller span of languages in their pre-training data. In fact, Bloomz and Aya show near-random performance on date-recall tasks related to wars and leaders. The performance variation across languages for the same country is further analyzed using standard deviation in Appendix~\ref{sec:eval_variation_across_languages_appendex}.

\subsection{Western vs. Non-Western Languages} 
Our analysis reveals a significant performance gap between Western (English, German, Spanish, Russian) and non-Western (Hindi, Japanese, Chinese) languages, as shown in Table~\ref{tab:main_results}. Models are consistently more culturally literate when queried in Western languages, a finding that is statistically significant (see Table~\ref{tab:p_value} in Appendix for the statistical test results). While the composition of pre-training data partially explains this trend, it does not account for all anomalies. For instance, the underperformance of the LLaMA-2 series in Hindi and that of the Mixtral series in Hindi, Japanese, and Chinese is expected, as these languages are underrepresented in their respective training data. However, some results are counterintuitive. Mixtral, for example, despite being primarily trained on English, German, and Spanish, performs substantially better in Russian than in other non-primary languages like Hindi, Japanese or Chinese. Similarly, Russian outperforms Hindi for LLaMA-3, even though the model is purportedly optimized more for Hindi. These discrepancies, alongside GPT-4's varied performance across Western and non-Western languages for certain domains, and the sub-par performance of open-source models in non-Western languages, raise concerns about their inclusivity for a diverse global user base~\citep{Blasi2021SystematicII}.

\subsection{Transferability of Cultural Knowledge Across Languages}
While the previous analysis focused on accuracy in individual languages, a crucial aspect of true multilingual proficiency is knowledge transferability. For a model to be considered culturally literate, it should be able to answer factual questions consistently, regardless of the language used. To quantify this, we introduce two novel metrics: \textit{Total Coverage} and \textit{Smooth Coverage}.

\paragraph{Total Coverage (TC).} This metric evaluates the consistency of complete factual recall. To measure how consistently a model knows a specific fact across multiple languages, we first define \textit{coverage} ($C^d_l$) for a given language $l$ and domain $d$ as the set of all entities in $d$ for which the model correctly answers at least three of the four associated questions in the given language. Then, similar to \citet{Qi2023CrossLingualCO}, we define \textit{Total Coverage} ($TC^d$) as the ratio of entities covered across \textit{all} tested languages to the total number of entities covered in \textit{any} language:
\begin{equation}
    TC^d= \frac{\mid \bigcap_{l \in L} C^d_l \mid}{\mid \bigcup_{l \in L} C^d_l \mid}
\end{equation}
$TC^d$ directly measures a model's ability to transfer knowledge. A high $TC^d$ score indicates strong knowledge transference, even if overall accuracy is modest. We evaluate $TC^d$ under four distinct scenarios to probe different aspects of this transferability:
\begin{enumerate}[nosep]
    \item[(i)] \textbf{Total Coverage (All),  $TC^d(All)$.} This measures cultural transferability across all seven languages in \datasetname, providing a holistic view of a model's multilingual alignment on our dataset.
    \item[(ii)] \textbf{Total Coverage (Pre-training data), $TC^d(Pre-Train)$.} To provide a fairer assessment of models like LLaMA and Mistral, where some languages are known to be underrepresented in their training data, this metric calculates TC only on the languages well-represented in each model's pre-training corpus.
    \item[(iii)] \textbf{Total Coverage (West), $TC^d(W)$.} Given the observation from Table~\ref{tab:main_results} that models perform better in Western languages, this metric quantifies their knowledge transferability specifically within this language group (English, German, Spanish, Russian).
    \item[(iv)] \textbf{Total Coverage (English Non-Western), $TC^d(Eng-NW)$.} Since English is the dominant language for all models, this scenario measures how well knowledge transfers from English to the non-Western languages in our dataset.
\end{enumerate}

The results, presented in Table~\ref{tab:TC}, reveal a significant lack of knowledge transfer in most models. The $TC(All)$ score for open-source models is notably low, often near $5\%$, indicating that a vast majority of their cultural knowledge is not consistently accessible across languages. While the $TC(Pre-Train)$ scores show an expected increase (averaging 10.4$\times$), the gap remains substantial. We also observe a strong bias towards Western languages~\citep{Naous2023HavingBA}; for instance, Mixtral and Mistral show a nearly $30\times$ and $18\times$ jump in transference for $TC(W)$, respectively, compared to $TC(All)$. This jump can be attributed to the language distribution in their pre-training data, as we saw a similar jump when comparing $TC (All)$ and $TC(Pre-Train)$ for these models.  Interestingly, other models show stronger transference for $TC(W)$ than $TC(Pre-Train)$ indicating a bias towards Western languages, as also discussed in earlier section. The consistently low $TC(Eng-NW)$ scores further highlight the poor alignment between English and non-Western languages. Our pairwise analysis in Appendix~\ref{sec:apendix_pairwise} corroborates this, showing that Western language pairs have significantly higher TC scores.

\begin{table}[!t]
\centering
\adjustbox{max width=\linewidth}{
\begin{tabular}{lcccc}
\hline
 & TC$^d$ &TC$^d$ & TC$^d$ & TC$^d$ \\ 
Model & (All) & (Pre-Train) & (W) & (Eng-NW) \\ \hline
\multicolumn{5}{c}{\cellcolor[HTML]{EEEEEE}Monuments} \\ 
\hline
Bloomz-7B1 & 0.00&0.00&6.08&0.89\\
LLaMA-2-7B-Chat & 0.38&6.51&26.46&0.82\\
Mistral-7B-Instruct & 6.34&78.16&57.20&7.34\\
Meta-Llama-3-8B-Instruct & 14.18&30.26&34.59&18.95\\
LLaMA-2-13B-Chat & 2.27&8.71&44.87&2.40\\
Aya & 6.01&6.01&21.90&10.70\\
GPT-4 & 62.59&62.59&81.88&72.56\\
Mixtral-8x7B & 2.16&83.33&72.43&2.59\\
\hline
\multicolumn{5}{c}{\cellcolor[HTML]{EEEEEE}Leaders} \\ 
\hline
Bloomz-7B1 & 3.75&3.75&9.84&9.70\\
LLaMA-2-7B-Chat & 0.48&3.40&6.95&1.05\\
Mistral-7B-Instruct & 2.29&54.62&41.54&4.03\\
Meta-Llama-3-8B-Instruct & 10.25&44.80&48.75&12.17\\
LLaMA-2-13B-Chat & 6.67&14.12&24.71&8.28\\
Aya & 16.95&16.95&30.69&22.45\\
GPT-4 & 28.46&28.46&50.83&33.48\\
Mixtral-8x7B & 2.46&86.59&73.05&2.57\\
\hline
\multicolumn{5}{c}{\cellcolor[HTML]{EEEEEE}Wars} \\ 
\hline
Bloomz-7B1 & 5.21&5.21&11.39&13.50\\
LLaMA-2-7B-Chat & 3.47&9.31&16.67&8.53\\
Mistral-7B-Instruct  & 3.68&42.25&35.40&5.29\\
Meta-Llama-3-8B-Instruct  & 21.22&37.09&38.81&27.81\\
LLaMA-2-13B-Chat & 2.75&12.05&18.16&5.07\\
Aya & 5.28&5.28&10.79&10.77\\
GPT-4 & 47.23&47.23&70.76&52.36\\
Mixtral-8x7B & 9.45&71.26&59.33&11.30\\
\hline
\multicolumn{5}{c}{\cellcolor[HTML]{EEEEEE}National Parks} \\ 
\hline
Bloomz-7B1 & 0.29&0.29&7.72&0.65\\
LLaMA-2-7B-Chat & 0.48&5.60&21.54&0.52\\
Mistral-7B-Instruct & 1.41&67.08&43.45&1.78\\
Meta-Llama-3-8B-Instruct & 7.94&27.74&31.54&11.45\\
LLaMA-2-13B-Chat & 0.49&6.55&34.25&0.53\\
Aya & 5.17&5.17&25.17&8.39\\
GPT-4 & 44.35&44.35&86.42&47.76\\
Mixtral-8x7B & 1.08&77.88&60.35&1.11\\

 \hline

\end{tabular}
}
\caption{Total Coverage ($TC$) scores across the four domains, evaluated under different language scenarios. $TC(All)$ shows overall transferability, which is low for most open-source models. Scores improve for scenarios limited to pre-training languages ($TC(Pre-Train)$) and Western languages ($TC(W)$), indicating a strong data bias. GPT-4 consistently outperform other models, especially in the all-language scenario.
}
\label{tab:TC}
\vspace{-5mm}
\end{table}

\paragraph{Smooth Coverage (SC).} The binary nature of TC (an entity is either covered or not) can be overly strict, as it fails to credit models for partial knowledge (e.g., answering 2 of 4 questions correctly). To address this, we introduce \textit{Smooth Coverage (SC)}, a more nuanced metric that uses a fuzzy set extension. First, for each entity $e$ in the domain $d$, we define its membership score, $m^{(e)}_l$, as the proportion of correctly answered questions for that entity in language $l$:
\begin{equation}
    m^{(e)}_l = \frac{1}{P}\sum_{p=1}^{P}{\mathds{I}(y_{pred}=y_{truth})_{l,p}^{(e)}}
\end{equation}
where, $P$ is the total number of prompts for a question on the entity $e$, $y_{pred}$ and $y_{truth}$ are the model generated and ground-truth answers respectively, and $\mathds{I}(.)$ is the indicator function. This score reflects the degree of knowledge and converts $C^d_l$ into a fuzzy set. We then define $SC$ as the average ratio of the minimum membership score to the maximum score for each entity across all languages:
\begin{equation}
    SC^d_l = \frac{1}{|\mathcal{E}|} \sum_{e \in \mathcal{E}} \frac{\min_{l \in L} m^{(e)}_l}{\max_{l \in L} m^{(e)}_l + \epsilon}
\end{equation}
where, $\mathcal{E}$ is the set of entities in domain $d$, and $\epsilon$ is a small constant to prevent division by zero. This metric rewards models for consistent, even if partial, knowledge.

\begin{table}[ht]
\centering
\adjustbox{max width=\linewidth}{
\begin{tabular}{l|cccc}
\toprule
\textbf{Model} & \textbf{M} & \textbf{L} & \textbf{W} & \textbf{N} \\
\midrule
Bloomz-7B1              & 0.71  & 3.66  & 5.03  & 0.84  \\
LLaMA-2-7B-Chat         & 1.24  & 1.33  & 7.50  & 1.64  \\
Mistral-7B-Instruct     & 6.94  & 2.58  & 5.56  & 1.95  \\
Meta-LLaMA-3-8B-Instruct   & 20.13 & 14.16 & 24.11 & 12.06 \\
LLaMA-2-13B-Chat        & 4.27  & 7.33  & 4.68  & 2.59  \\
Aya                    & 7.02  & 7.83  & 4.47  & 5.34  \\
GPT-4          & 61.92 & 23.99 & 39.94 & 43.97 \\
Mixtral-8x7B           & 5.07  & 3.75  & 13.27 & 2.48  \\
\bottomrule
\end{tabular}
}
\caption{Smooth Coverage (SC) scores across all languages and domains. GPT-4 demonstrates the most robust cross-lingual transference. In contrast, Bloomz, despite its linguistically diverse training, shows poor transference across languages for all domains (\textbf{M}: Monuments, \textbf{L}: Leader,\textbf{ W}: War, and \textbf{N}: National Park).
}
\label{tab:smooth_coverage}
\end{table}

As reported in Table~\ref{tab:smooth_coverage}, the SC scores confirm the trends observed with TC. While the scores are numerically higher because they account for partial knowledge, the overall conclusion remains unchanged. GPT-4 demonstrates the strongest cross-lingual robustness, while models like Bloomz, despite their linguistically diverse training data, score the lowest. This finding reinforces that current open-source models are still largely inadequate at providing faithful and consistent answers in multilingual scenarios.

\begin{figure}[!t]
\begin{center}
\includegraphics[width=\linewidth]{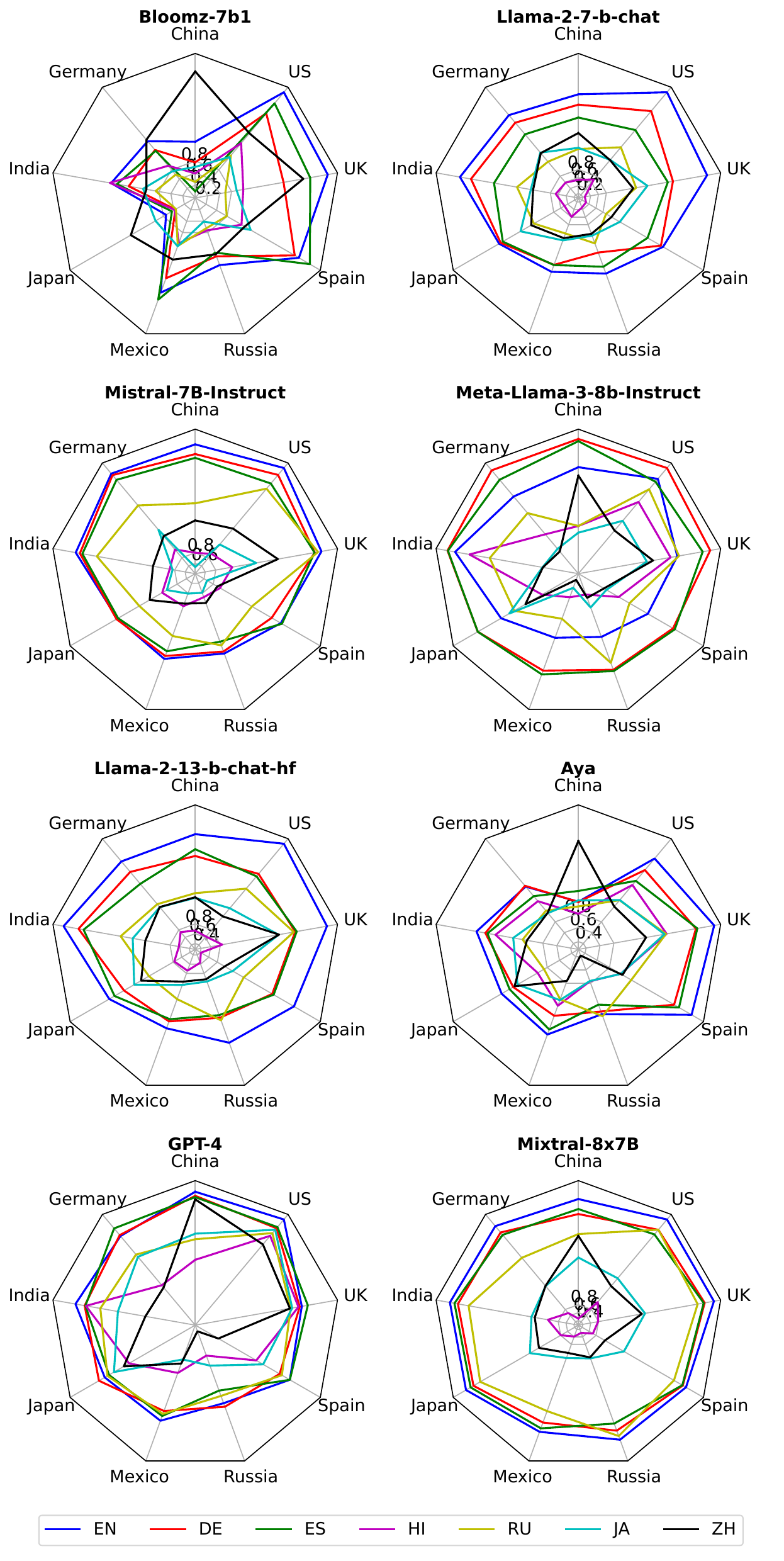}
\caption{A detailed breakdown of model accuracy across different languages for each of the nine countries in \datasetname~(see Table~\ref{tab:langs_iso} for ISO codes). Each radar plot represents a model, with axes for the nine countries and colored lines for the seven languages. The plots demonstrate the significant performance disparities across both nations and languages. For instance, most models show stronger performance in English (blue line) across all countries, while struggling with non-Western languages like Hindi (magenta line). The irregular shapes of the plots for models like Bloomz or LLaMA-2 highlight inconsistent cultural knowledge across different nations.
}
\label{fig:combined}
\end{center}
\vspace{-5mm}
\end{figure}

\subsection{Disparities in Cultural Literacy Across Nations}
While we previously analyzed the performance of the models aggregated by language, let us now look into how cultural literacy varies across different nations, revealing significant inconsistencies. As visualized in the radar plots in Figure~\ref{fig:combined}, no single model demonstrates uniform cultural knowledge across all nine countries. The asymmetrical shapes of the plots clearly indicate that a model's proficiency in one nation's cultural facts does not guarantee similar proficiency in another's, even when queried in the same language.

\paragraph{Performance on Leaders.}
A topic-by-topic analysis reveals further nuance. For questions about the birth year of \textit{leaders} (see Appendix Figure~\ref{fig:leader_country}), Mixtral-8x7B is the most consistent performer across different nations. However, this consistency is largely confined to Western languages; its performance drops significantly in non-Western languages, a trend observed across all models. LLaMA-3-8B-Instruct also shows reasonable consistency, particularly in English, German, and Spanish, though it notably struggles with facts about Mexico.

\paragraph{Performance on Monuments.} When recalling the location of \textit{monuments} (c.f. Appendix Figure~\ref{fig:monuments_country}), most models again perform well in Western languages. The exceptions are Aya and Bloomz, which struggle more broadly. The LLaMA-2 family exhibits a specific weakness, showing low performance on facts related to Japan, Mexico, and China, even when queried in those nations' dominant languages. Among all models, only GPT-4 maintains a relatively uniform level of literacy across both Western and non-Western languages like Japanese and Hindi for this topic.

\paragraph{Performance on National Parks and Wars.}
The highest degree of variability is observed for questions about \textit{national parks} (Appendix Figure~\ref{fig:national_park_country}) and \textit{wars} (Appendix Figure~\ref{fig:war_country}). For these topics, performance is highly inconsistent, fluctuating significantly between both nations and languages for nearly all models. The only notable exception is GPT-4's relatively stable and high performance on national park locations when queried in English, German, and Spanish.

\subsection{Cultural Knowledge vs. Linguistic Competence}
Finally, we analyze model performance averaged across all languages to see how culturally literate models are about different nations overall (see Appendix Figure~\ref{fig:country_performance}). This perspective reveals a surprising trend that contrasts with our earlier language-based findings.

\paragraph{Performance by Nation.}
When viewed by country, the performance hierarchy is largely consistent with previous results: GPT-4 exhibits the strongest cultural literacy (nearly $80\%$ accuracy for the US, China, and India), followed by LLaMA-3-8B-Instruct and Mixtral-8x7B (above $60\%$ for the same countries). The LLaMA-2 series remains in the moderate $40-60\%$ range, while Aya and Bloomz show the lowest literacy, often below $40\%$ for most countries.

\paragraph{Decoupling of Knowledge and Language.}
Most interestingly, the strong performance divide between Western and non-Western languages does \textit{not} translate into a similar divide between Western and non-Western \textit{nations}. As shown in Table~\ref{tab:main_results}, models often demonstrate deeper cultural knowledge of countries like China, India, and Japan than of Western nations like Germany or Spain. This occurs even while the models perform poorly in the native languages of those non-Western nations (e.g., Chinese, Hindi). This decoupling of linguistic competence from cultural knowledge is a key finding of our study. It suggests that a model can be highly knowledgeable about a particular nation’s facts, even if it cannot express that knowledge effectively in that nation’s primary language.

\section{Conclusion}
In this work, we introduced \datasetname, a large-scale multilingual benchmark designed to evaluate the cultural literacy of LLMs beyond the typical Western-centric scope. Our analysis of eight models revealed significant inconsistencies in their factual knowledge across languages and nations. We found a strong performance bias towards Western languages, though this did not always translate to better knowledge of Western countries. Models specifically designed for broad language support, like Aya and Bloomz, struggled with factual recall, particularly with dates. Furthermore, our novel transference metrics showed that open-source models have a severe limitation in transferring cultural knowledge across languages, highlighting a critical gap between them and the proprietary LLMs. These findings underscore the need for more culturally inclusive training and evaluation methods to create truly global and equitable language models.

\section*{Limitations}

This work attempts to present a more inclusive approach towards benchmarking the cultural literacy of models across nations and demonstrates the disparity in LLMs' performance with variation in language. However, it has three main limitations. First, due to computational constraints, we could not experiment with larger multilingual LLMs. Consequently, while our work benchmarks a range of widely-used models, the cultural literacy of colossal models remains an important direction for future investigation.

Second, the design of \datasetname~has limitations in scope and structure. Our selection of nations based on widely spoken languages resulted in no coverage of countries from Africa and South America. Furthermore, while the dataset is parallel across languages, the number of factual entities for each domain (e.g., wars, monuments, etc.) naturally varies between nations, an inherent constraint of nation-specific factual benchmarking approach.

Finally, this study's scope is focused on performance benchmarking rather than an in-depth analysis of the models' training data. Such an analysis could provide insights into our findings, such as why models like Bloomz and Aya exhibit lower cultural literacy despite their linguistically inclusive training. Investigating the topic distribution and biases in their publicly available training corpora, like that of BLOOM~\citep{laurenccon2022bigscience}, is a valuable avenue for future research.

\section*{Ethical Statement}
This work aims to evaluate the cultural literacy of LLMs across multiple languages and nations by probing their knowledge of historically and culturally significant facts. We take care to ensure that our dataset is balanced, factual, and respectful of diverse cultures, avoiding stereotypes or biased representations. All cultural content is sourced from publicly available, reputable references such as encyclopedias, official historical records, and recognized heritage listings.

We acknowledge that cultural knowledge is complex and dynamic, and our work does not attempt to capture the full richness of any culture or community. Instead, it focuses on well-documented factual knowledge as a proxy for measuring models’ cross-cultural understanding. We encourage future research to incorporate a wider range of cultural perspectives and to evaluate the social impact of deploying LLMs in multicultural contexts.

\section*{Acknowledgements}
Anwoy Chatterjee gratefully acknowledges the support of the Google PhD Fellowship. Tanmoy Chakraborty acknowledges the support of the Anusandhan National Research
Foundation (DST/INT/USA/NSF-DST/Tanmoy/P-2/2024) and Rajiv Khemani Young Faculty
Chair Professorship in Artificial Intelligence. The authors acknowledge the support
of Google GCP Grant.

\bibliography{custom}

\appendix

\section{Appendix}
\label{sec:appendix}
\subsection{Prompt Templates}
\label{sec:prompt_tempt}
Table~\ref{tab:examples} shows a few question templates used in our dataset creation. We translate the question template along with the entity-answer tuples and options to create \datasetname.

\subsection{ISO Codes} \label{appendix:iso}
Table~\ref{tab:langs_iso} contains the ISO codes of the seven languages used in our dataset.

\begin{table}[!ht]
\small
    \centering
    \begin{tabular}{ccc}
    \toprule
        \textbf{Language} & \textbf{ISO 639-1 code} & \textbf{Family} \\
        \midrule
        English & EN &  IE: Germanic \\
        \midrule
        German & DE & IE: Germanic \\
        \midrule
        Hindi & HI &  IE: Indo-Iranian \\
        \midrule
        Chinese & ZH &  Sino-Tibetan \\
        \midrule
        Russian & RU &  IE: Balto-Slavic \\
        \midrule
        Spanish & ES & IE: Italic  \\
        \midrule
        Japanese & ZH & 	Japonic  \\
        \midrule
    \bottomrule
    \end{tabular}
    \caption{List of languages and their ISO codes used in our experiments.}
    \label{tab:langs_iso}
\end{table}

\subsection{Selecting Languages and Entity Distribution}
\label{sec:dataset_dis}
\begin{figure}[!ht]
\begin{center}
\includegraphics[width=0.8\linewidth]{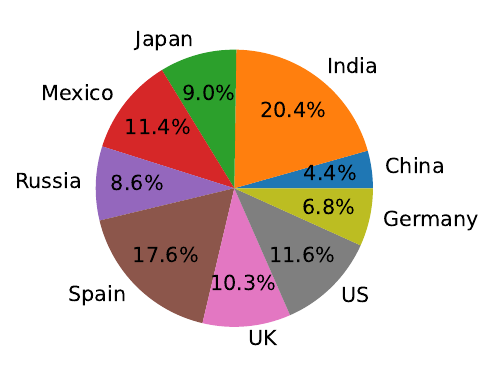}
\caption{Distribution of entities across the nine nations considered in our study.}
\label{fig:pie}
\end{center}
\vspace{-5mm}
\end{figure}

We started our study by selecting the three most widely spoken languages in the world (Mandarin, Spanish and English). One nation from each of these languages (China, Spain, US) was selected in the beginning. We then expanded our selection to include a combination of diverse (Hindi, Russian) and similar languages (German and Japanese), with one nation where they are widely spoken (India, Russia, Germany). This resulted in a set of seven nations and languages. We further included Mexico and UK to expand on the cultures we cover. Figure~\ref{fig:pie} shows the distribution of entities over the nine countries.

\subsection{Pre-Training Datasets of the LLMs Used in Our Study}
\label{sec:model_language_apendix}

\begin{enumerate}
    \item[(i)] \textbf{Aya:} Aya is an instruction-tuned mT5 model\citep{xue2020mt5} that supports $101$ languages, including all seven languages used in our study.
    \item[(ii)] \textbf{Bloomz:} Bloomz supports $46$ languages, including programming languages. It covers all seven languages used in our study.
    \item[(iii)] \textbf{Mistral:} The Mistral family of models primarily focuses on European languages, covering English, French, Italian, German, and Spanish. The exact distribution of its training dataset is unknown, so its coverage of other languages is unknown.
    \item[(iv)] \textbf{LLaMA-2:} LLaMA-2's pre-training dataset predominantly consists of English, with $89.7\%$ of the data in English. The remaining $8.38\%$ comprises unknown languages (mainly programming languages), and about $2\%$ consists of non-English languages. Except for Hindi, all other languages in our dataset seem to be represented.
    \item[(v)] \textbf{LLaMA-3:} LLaMA-3 builds upon the multilingual capabilities of LLaMA-2, with $8\%$ of its training dataset dedicated to multiple languages. Although LLaMA-3 has been trained on a wide array of languages, it is specifically optimized and safety-tuned for eight languages: English, German, French, Italian, Portuguese, Hindi, Spanish, and Thai.
    \item[(vi)] \textbf{GPT-4:} The exact details of its training dataset are not publicly available.
\end{enumerate}

\begin{table*}[!ht]
\small
\begin{center}
\scalebox{0.8}{
\begin{tabular}{ |m{2cm}|m{12.4cm}| m{2.8cm}|}
 \hline
 {\bf Domain}&{\bf Question Template}&{\bf Entity-Answer example} \\ \hline
Year of War&Question: As per American history, when did \{War\} happen?

\{Option\}

Answer: 

Question: What year did \{War\} start according to American history?

\{Option\}

Answer: 

Question: Historical accounts from American indicate \{War\} occur in?

\{Option\}

Answer:

Question: Can you tell me according to American history the date when \{War\} took place?

\{Option\}

Answer: & \textit{(American Revolutionary War, 1775)}\\

\hline

Location of national park or monument&Question: Where can \{Location\} be found within China?

\{Option\}

Answer: 

Question: In which part of China is \{Location\} situated?

\{Option\}

Answer: 

Question: Where in China is \{Location\} located?

\{Option\}

Answer: 
    
Question: In which area of China can \{Location\} be found?

\{Option\}

Answer: &\textit{(Wuyi Mountains National Park, Fujian)}\\
\hline
Birth year of Leader& 'Question: When was German leader \{NAME\} born?

\{Option\}

Answer: 

Question: What is the birthyear of the German leader \{NAME\}?

\{Option\}
    
Answer: 

Question: On what year was the German leader \{NAME\} born?

\{Option\}

Answer: 

Question: What is the year of birth of the German leader \{NAME\}?

\{Option\}

Answer: & \textit{(Olaf Scholz,1958)}\\
\hline
\end{tabular}
}
\caption{Question Template in English along with an example entity-answer. We create the option using three negative samples and the true answer to fill the \{Option\}.}
\vspace{-2mm}
\label{tab:examples}
\end{center}
\end{table*}

\begin{table}[!t]
\centering
\adjustbox{max width=1\linewidth}{
\begin{tabular}{lcccc}
\hline
Model & M & L& W & N\\ \hline
\hline
Bloom-7B1 & \textbf{1.9e-47}&0.38&\textbf{0.002}&\textbf{1.23e-59}\\
LLaMA-2-7B-chat & \textbf{2.82e-160}&\textbf{4.16e-39}&\textbf{1.20e-38}&\textbf{2.83e-195}\\
Mistral-7B-Instruct & \textbf{4.96e-113}&\textbf{1.324e-5}&\textbf{3.06e-243}&\textbf{3.77e-189}\\
Meta-LLaMA-3-8B & \textbf{1.145e-31}&\textbf{3.90e-86}&\textbf{1.82e-32}&\textbf{1.02e-81}\\
LLaMA-2-13B-chat & \textbf{1.83e-32}&\textbf{7.35e-15}&\textbf{9.82e-12}&\textbf{4.67e-193}\\
Aya & \textbf{3.73e-45}&0.45&0.02&\textbf{1.06e-21}\\
Mixtral-8x7B & \textbf{1.09e-132}&\textbf{2.67e-156}&\textbf{1.47e-105}&\textbf{3.99e-247}\\
GPT-4 &\textbf{1.25e-13}&\textbf{4.23e-9}&\textbf{2.45e-12}&\textbf{3.57e-51}\\
\hline
\end{tabular}
}

\caption{p-values for significance testing of the hypothesis that models perform better in Western languages. M (Monuments), L (Leader), W (War) and N (National Park). Significance is decided by taking $\alpha = 0.05$ and statistically significant results has been marked bold.}
\label{tab:p_value}
\vspace{-5mm}
\end{table}

\subsection{Pairwise Total Coverage Scores of Models}
\label{sec:apendix_pairwise}
\begin{figure*}[!ht]
\begin{center}
\includegraphics[width=\textwidth]{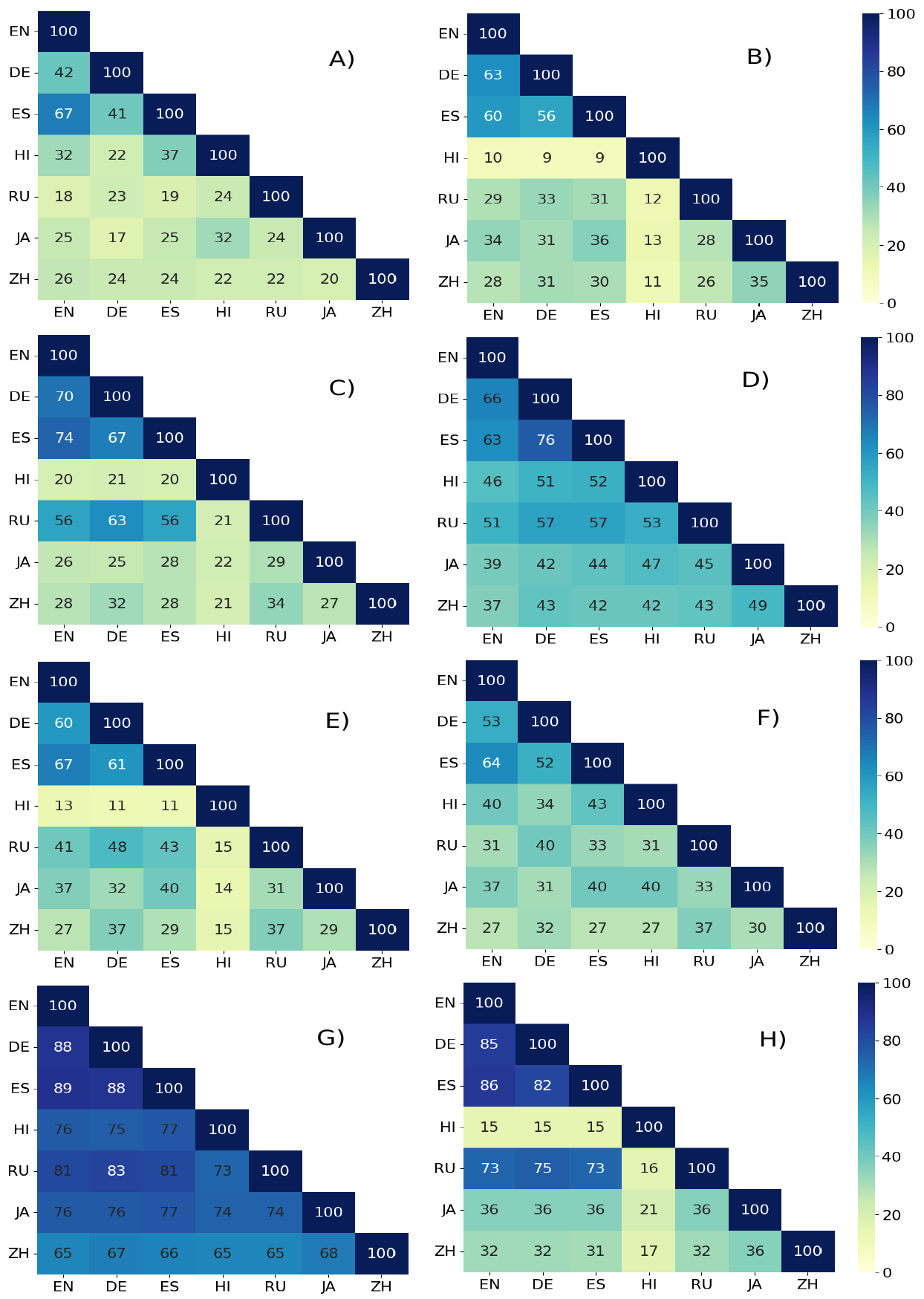}
\caption{Heatmaps for pairwise $TC$ of all language pairs, for -- (A) Bloomz-7B1, (B)LLaMA-2-7B-Chat, (C)Mistral-7B-Instruct, (D)Meta-LLaMA-3-8B-Instruct, (E) LLaMA-2-13B-Chat, (F) 13-billion Aya, (G) GPT-4 and (H) Mixtral-8x7B. 
}
\label{fig:total_pairwise_coverage}
\end{center}
\vspace{-5mm}
\end{figure*}

Figure~\ref{fig:total_pairwise_coverage} represents the pairwise $TC$ of Bloomz-7B1, LLaMA-2-7B-Chat, Mistral-7B-Instruct, LLaMA-2-13B-Chat, and Aya. We see that the model generally sees a better $TC$ score in Western languages compared to non-Western ones, highlighting that the models struggle in non-Western languages. Mistral-7B-Instruct and LLaMA-2-13B-Chat show relatively better performance among the models evaluated, but none as strong as GPT-4, Mixtral or Meta-LLaMA-3 that have been discussed in the main text.

LLaMA-2 series shows very poor $TC$ for all language pairs consisting of Hindi, indicating the lack of consistency and understanding in Hindi. Aya while showing low absolute performance (c.f. Table~\ref{tab:main_results}), still shows overall reasonable $TC$ across languages.

\subsection{Evaluating Disparity Across National Park and War Questions}
\label{sec:radar_analysis_apendix}

\begin{figure}[!t]
\begin{center}
\includegraphics[width=\linewidth]{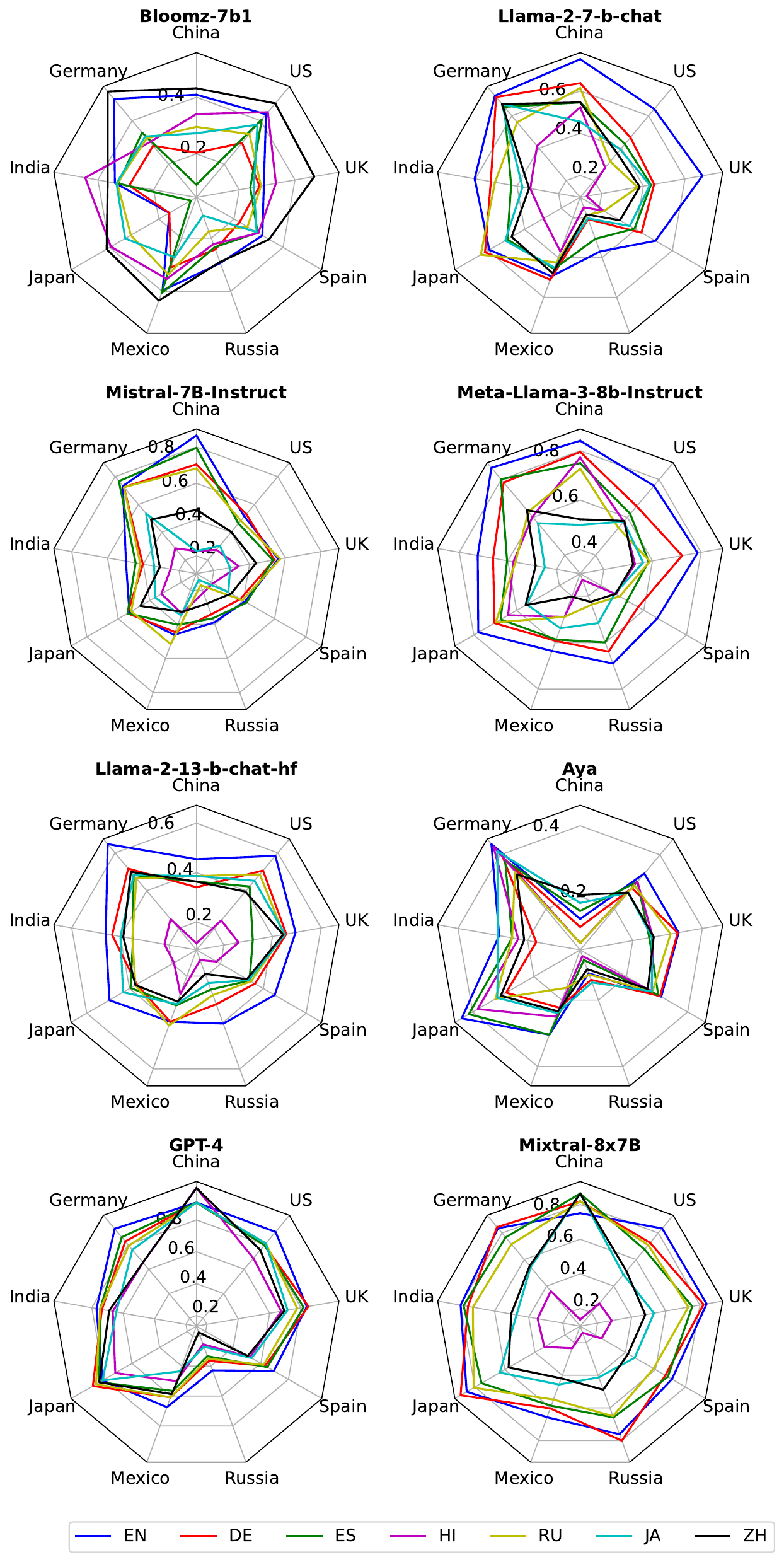}
\caption{Accuracy of models in different languages (indicated in different colours) for questions on the date of wars in different nations.
}
\label{fig:war_country}
\end{center}
\vspace{-5mm}
\end{figure}

Figure~\ref{fig:war_country} shows the performance of models on war-type questions. We note that models have different literacy on these facts with no model performing similarly across all countries or languages. Dates about wars Russia has taken part in seem to be difficult for the model to recall, except for Mixtral and Meta-LLaMA-3.

Figure~\ref{fig:heat_national_park} shows the performance of models on the location of national park-type questions. Here also we note that models tend to have different literacy on these facts with no model performing similarly. Except for GPT-4 for Western languages, there is no homogeneous performance across countries.

\begin{figure}[!t]
\begin{center}
\includegraphics[width=\linewidth]{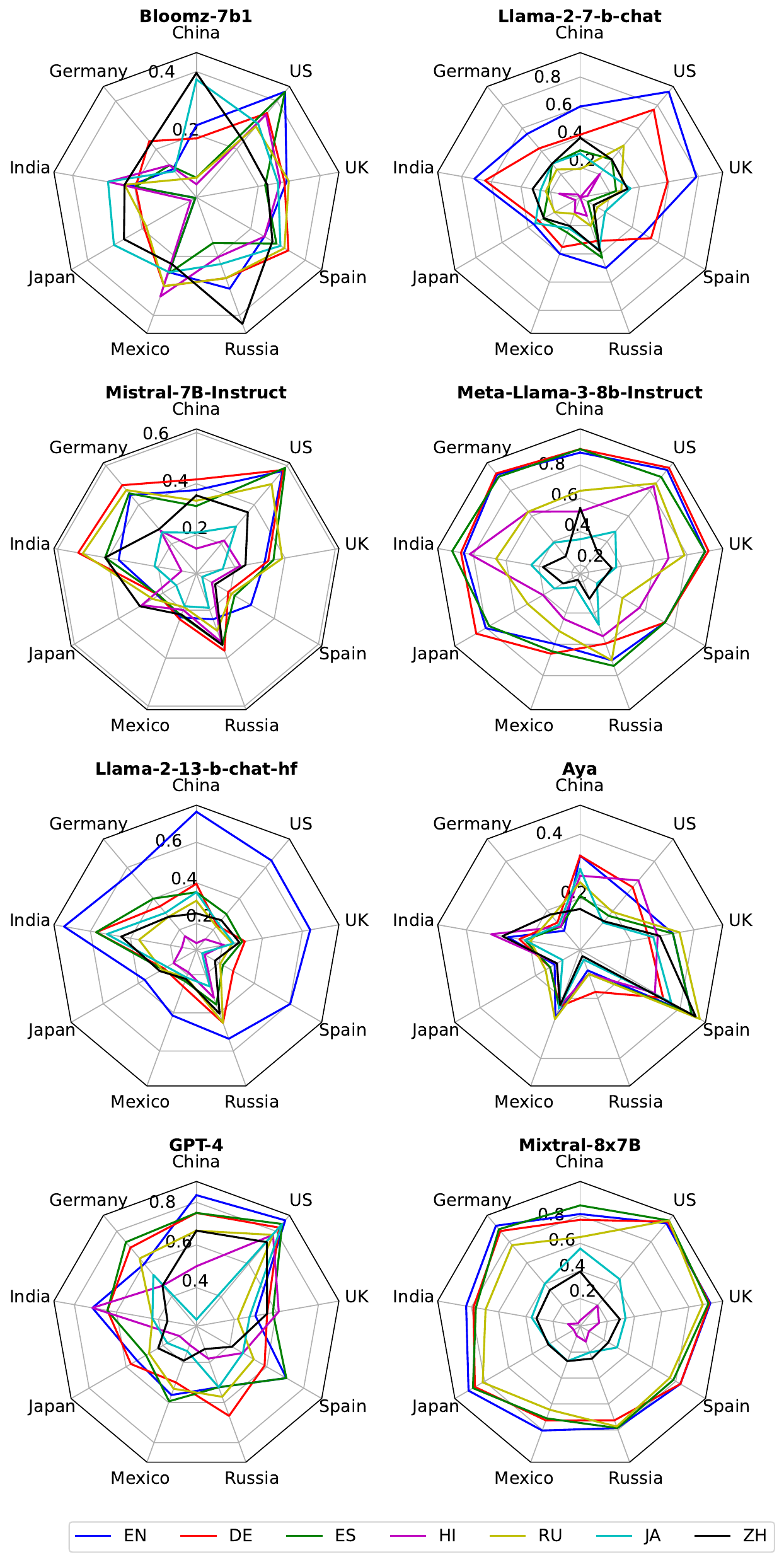}
\caption{Accuracy of models in different languages for questions on the birth year of leaders of different nations (see Table \ref{tab:langs_iso} for ISO code).
}
\label{fig:leader_country}
\end{center}
\vspace{-5mm}
\end{figure}

\begin{figure}[!t]
\begin{center}
\includegraphics[width=\linewidth]{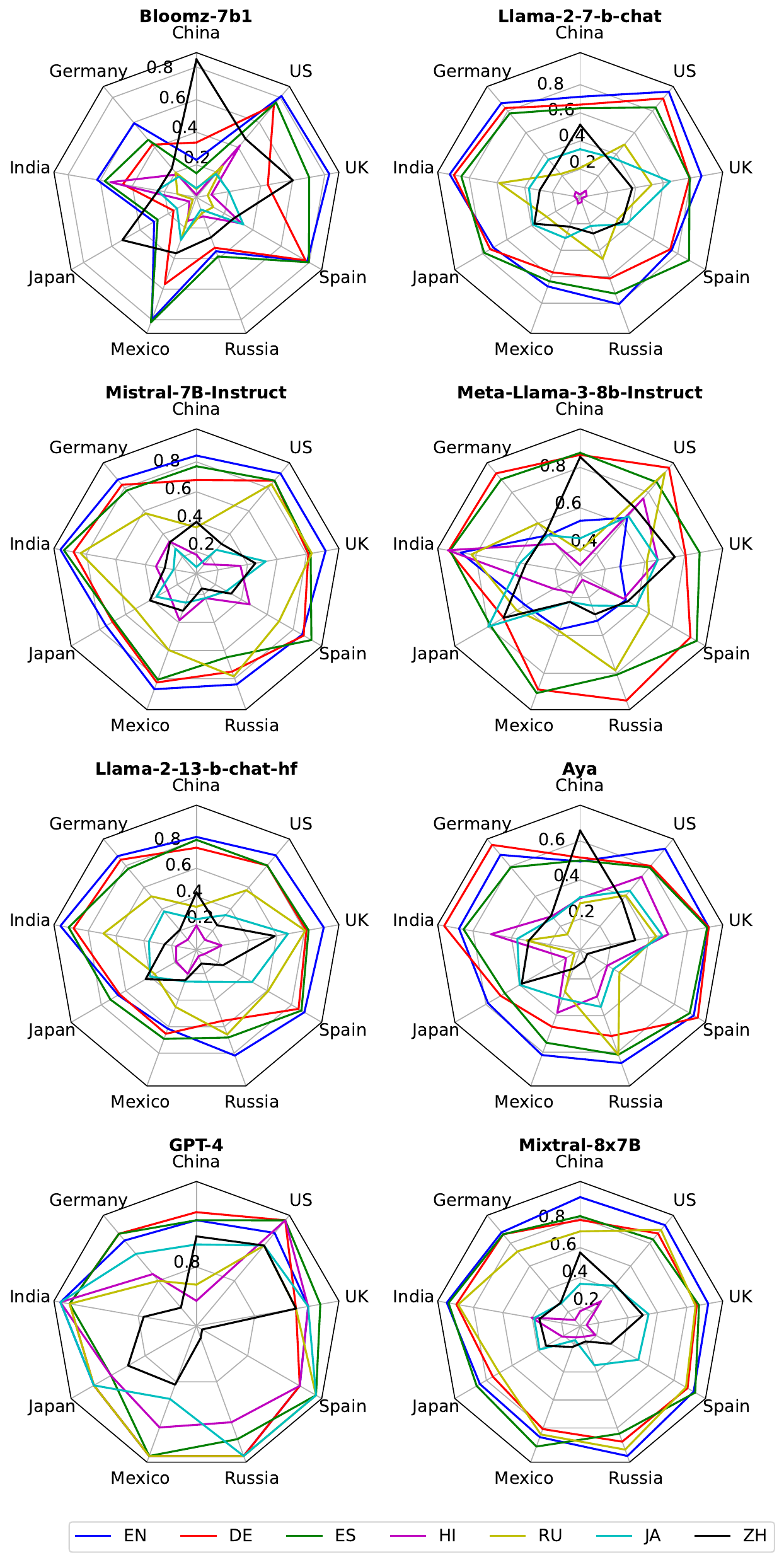}
\caption{Accuracy of models in different languages for questions on the location of UNESCO sites of different nations (see Table \ref{tab:langs_iso} for ISO code).
}
\label{fig:monuments_country}
\end{center}
\vspace{-7mm}
\end{figure}

\begin{figure}[!t]
\begin{center}
\includegraphics[width=\linewidth]{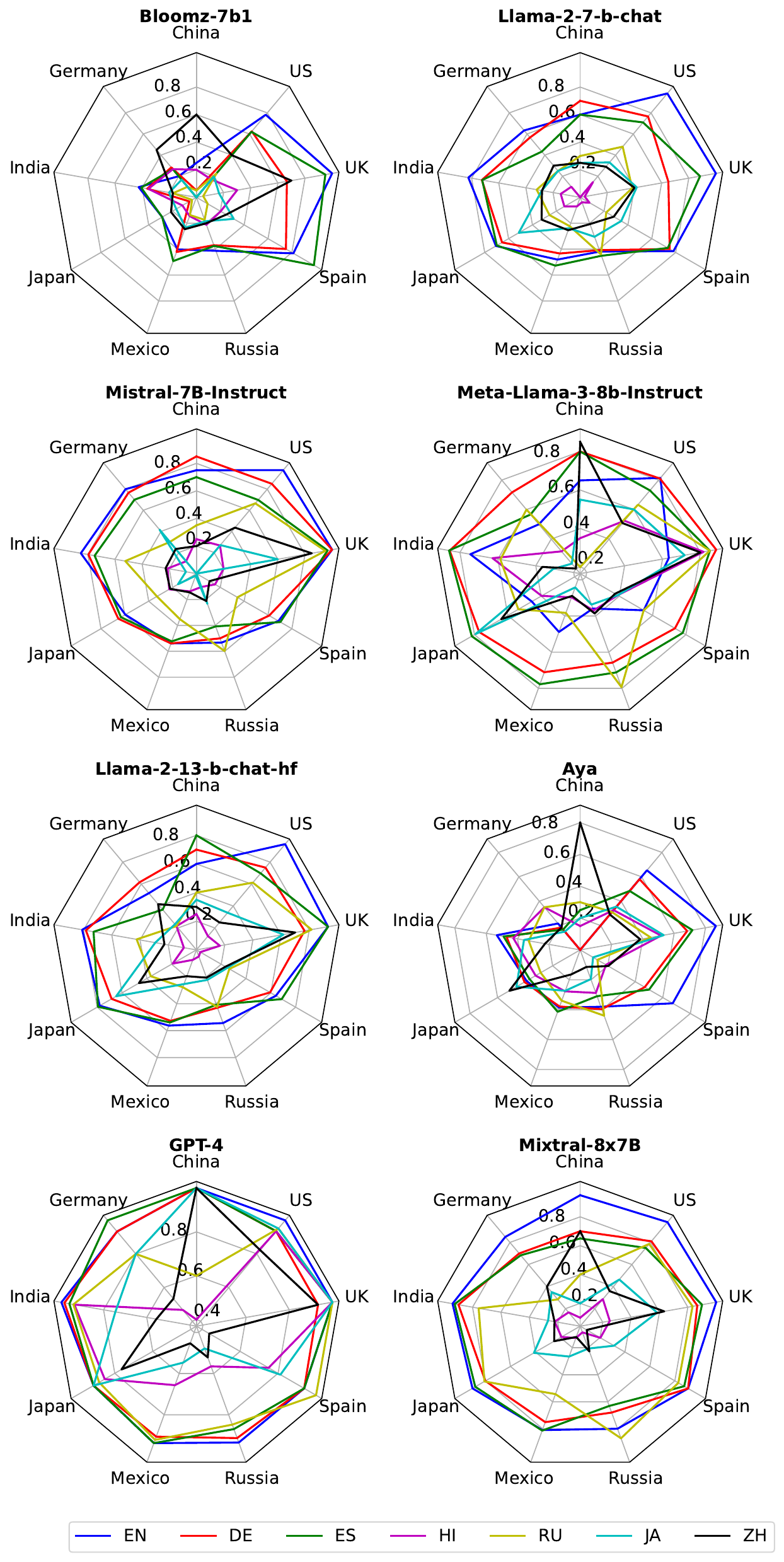}
\caption{Accuracy of models in different languages (indicated in different colours) for questions on the location of national parks of different nations.
}
\label{fig:national_park_country}
\end{center}
\vspace{-5mm}
\end{figure}

\subsection{Model-wise Analysis of Results} 
\label{sec:model_wise_analysis}
\begin{figure*}[!ht]
\begin{center}
\includegraphics[width=0.8\textwidth]{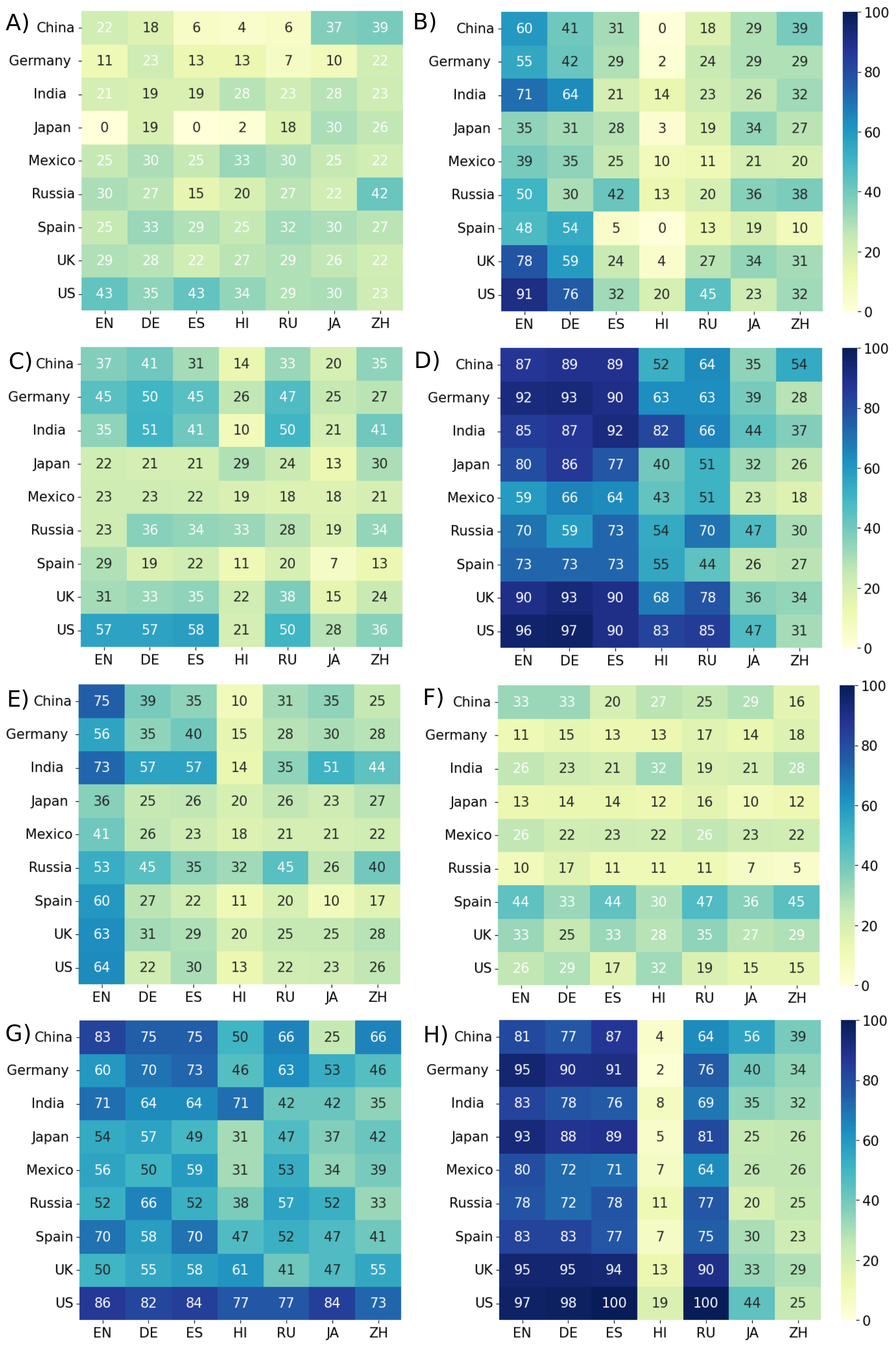}
\caption{Heatmap of accuracy on leader domain, for -- (A) Bloomz-7B1, (B) LLaMA-2-7B-Chat, (C) Mistral-7B-Instruct, (D) Meta-LLaMA-3-8B-Instruct, (E) LLaMA-2-13B-Chat, (F) 13-billion Aya, (G) GPT-4 and (H) Mixtral-8x7B. 
}
\label{fig:heat_leader}
\end{center}
\vspace{-5mm}
\end{figure*}

\begin{figure*}[!ht]
\begin{center}
\includegraphics[width=0.8\textwidth]{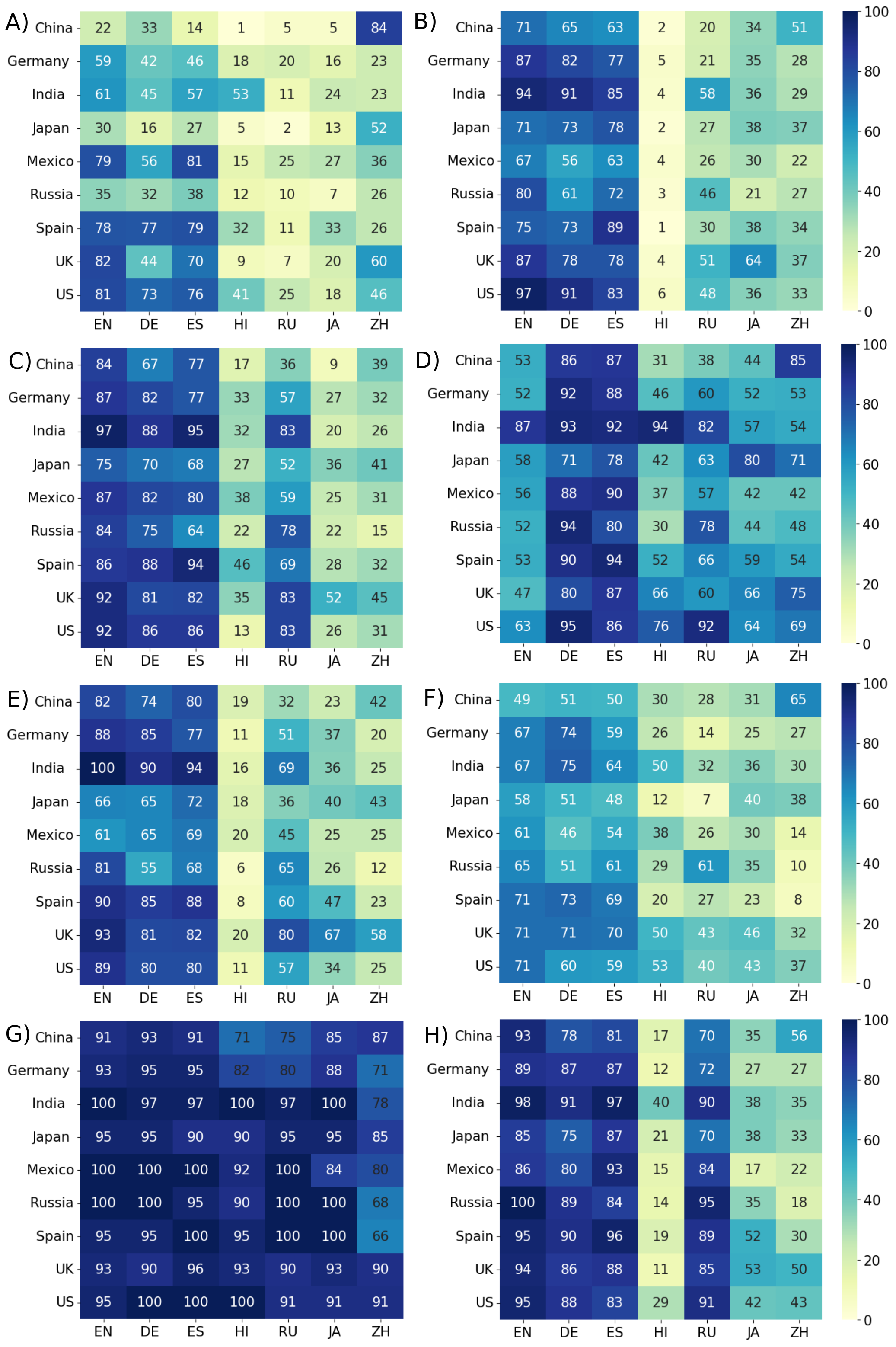}
\caption{Heatmap of accuracy on monuments domain, for -- (A) Bloomz-7B1, (B) LLaMA-2-7B-Chat, (C) Mistral-7B-Instruct, (D) Meta-LLaMA-3-8B-Instruct, (E) LLaMA-2-13B-Chat, (F) 13-billion Aya, (G) GPT-4 and (H) Mixtral-8x7B. 
}
\label{fig:heat_monuments}
\end{center}
\vspace{-5mm}
\end{figure*}

\begin{figure*}[!ht]
\begin{center}
\includegraphics[width=0.8\textwidth]{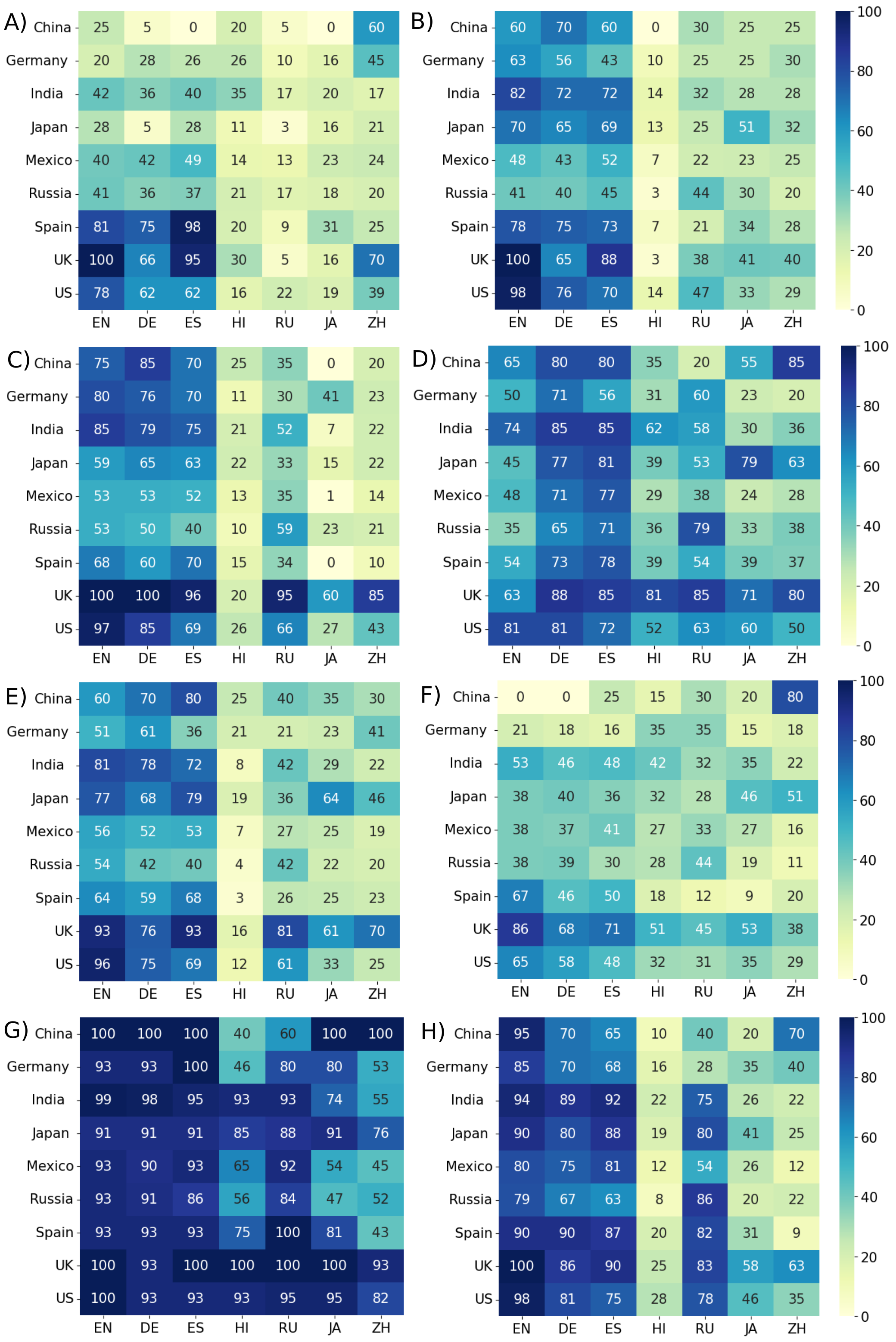}
\caption{Heatmap of accuracy on national park domain, for -- (A) Bloomz-7B1, (B) LLaMA-2-7B-Chat, (C) Mistral-7B-Instruct, (D) Meta-LLaMA-3-8B-Instruct, (E) LLaMA-2-13B-Chat, (F) 13-billion Aya, (G) GPT-4 and (H) Mixtral-8x7B. 
}
\label{fig:heat_national_park}
\end{center}
\vspace{-5mm}
\end{figure*}

\begin{figure*}[!ht]
\begin{center}
\includegraphics[width=0.8\textwidth]{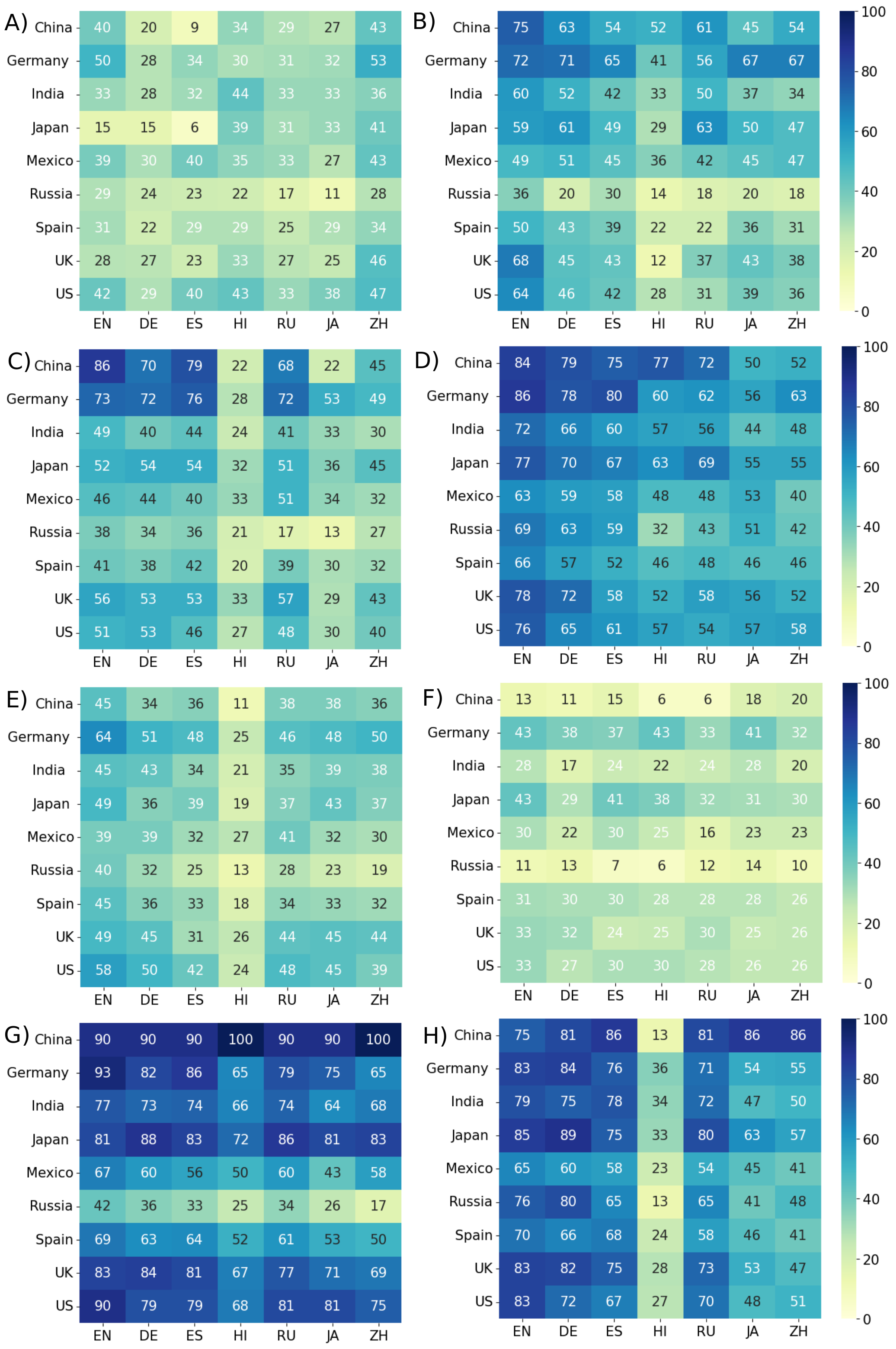}
\caption{Heatmap of accuracy on war domain, for -- (A) Bloomz-7B1, (B) LLaMA-2-7B-Chat, (C) Mistral-7B-Instruct, (D) Meta-LLaMA-3-8B-Instruct, (E) LLaMA-2-13B-Chat, (F) 13-billion Aya, (G) GPT-4 and (H) Mixtral-8x7B. 
}
\label{fig:heat_war}
\end{center}
\vspace{-5mm}
\end{figure*}

Here we dive deeper into analysing each model for a specific use case and along the country-language axes as shown in Figures~\ref{fig:heat_leader},~\ref{fig:heat_monuments},~\ref{fig:heat_national_park} and ~\ref{fig:heat_war}. The analysis is as follows:
\begin{enumerate}
    \item[(i)] \textbf{Bloomz-7B1:} We note that for war and leader domain questions, the model shows modest to poor performance in most cases, even worse than random in some instances. Despite being trained on a diverse multilingual corpus, Bloomz struggles with \datasetname~compared to LLaMA or Mistral models of the same size. For questions related to monuments and national parks, the model performs better in English, German, Spanish, and Chinese but shows low performance in other languages.
    
    \item[(ii)] \textbf{LLaMA-2-7B-Chat:} We consistently see that in all domains the model performs badly in the Hindi language (with accuracy lower than $20\%$). It seems to be most literate in Western-European languages which consistently outperform other languages for all countries and domains. Its coverage in Russian is still moderate compared to Western European languages. While it does show a decent level of literacy about a nation in the nation's native language, they are still not the best language to prompt the model in.
    
    \item[(iii)] \textbf{Mistral-7B-Instruct:} It shows modest performance on leader domain questions, but for other domains, we note it to be culturally literate. However, it is biased towards Western languages and shows only modest performance in other languages, especially Hindi.
    
    \item[(iv)] \textbf{Meta-LLaMA-3-8B-Instruct:} It is the only model below the size of 10 billion to be literate to some degree in all the languages covered by us, even Hindi. It still seems to be more proficient in Western languages than others.
    
    \item[(v)] \textbf{LLaMA-2-13B-Chat:} Similar to the 7-billion variant of LLaMA-2, this model also struggles to perform well in Hindi and has considerable disparity between Western-European languages and other languages.
    
    \item[(vi)] \textbf{Aya:} Like Bloomz, Aya has been trained on a diverse multilingual corpus, and similar to Bloomz it also struggles to perform well in war and leader domain questions. Its performance in other domains is better, but still not comparable to LLaMA-2-13B-Chat which is of a similar size.
    
    \item[(vii)] \textbf{GPT-4:} It is the most culturally literate model in multilingual setup, with considerably higher performance than all other models. There is an interesting observation though -- it struggles with questions about Germany if asked in Russian, especially for the monuments and national parks domains.
\end{enumerate}

\section{Human Evaluation of Translation Quality in \datasetname} 
\label{sec:HE}

To evaluate the translation quality of queries in our dataset, we recruited university students aged between $20$ to $25$ who were either native speakers of the language or had cleared language proficiency tests. We recruited a total of six language experts, one for each language. They were provided with the following Annotation Guidelines and Scoring Criteria.

\paragraph{Annotation Guidelines:} Each translation is evaluated based on the following criteria:
\begin{itemize}
    \item \textbf{Grammar} – Proper use of verbs, tense, agreement, and punctuation.
    \item \textbf{Fluency} – The sentence should sound natural and idiomatic.
    \item \textbf{Cohesion} – The sentence should be logically structured and clear.
\end{itemize}
\paragraph{Scoring Criteria:} Each translation is rated on a scale from $1$ to $5$:
\begin{itemize}
    \item \textbf{5 (Excellent)} – No grammatical errors. The sentence follows correct conjugations and agreements. It is natural, idiomatic, and fluent, resembling native speech.
    \item \textbf{4 (Good)} – Minor grammatical mistakes (e.g., incorrect prepositions, small conjugation errors), but they do not affect understanding. The sentence is still readable and sounds natural.
    \item \textbf{3 (Acceptable)} – Noticeable grammar mistakes, such as verb tense inconsistencies or incorrect word order. The sentence is understandable but slightly unnatural.
    \item \textbf{2 (Poor)} – Multiple grammatical mistakes significantly affecting readability. The sentence sounds unnatural or contains awkward phrasing.
    \item \textbf{1 (Unacceptable)} – The sentence is not grammatically correct and is difficult or impossible to understand. Severe errors break fluency, making the translation unusable.
\end{itemize}

\section{Evaluating the Variation of Model Performance Across Languages}
\label{sec:eval_variation_across_languages_appendex}

To study the disparity in a model's performance between languages on facts about the same country, we analyze the standard deviation of the accuracy of the models across different languages. A lower standard deviation indicates better homogeneous performance across languages. However, we observe high variance (c.f. Table~\ref{tab:sd}), indicating that multilingual LLMs struggle to transfer their cultural literacy across languages. In many instances, the standard deviation exceeds $20$, indicating a severe disparity in performance transference across languages.


\begin{table*}[!t]
\centering
\resizebox{1\textwidth}{!}{ 
\begin{tabular}{lccccccccc}
\hline
 \backslashbox{Model}{Country}&US&China&India&UK&Japan&Germany&Russia&Mexico&Spain \\ \hline
\multicolumn{10}{c}{\cellcolor[HTML]{EEEEEE}Monuments}     
 \\ \hline
Bloomz-7b1 &\color{red}$51.93\pm23.55$&$\color{red}24.05\pm26.89$&$\color{red}39.80\pm18.07$&$\color{red}42.08\pm28.19$&$\color{red}21.07\pm15.99$&$\color{red}32.30\pm15.57$&$\color{red}23.38\pm11.96$&$\color{red}46.29\pm24.76$&$\color{red}48.47\pm26.86$\\
LLaMA-2-7B-chat&$\color{red}56.85\pm32.05$&$\color{red}44.17\pm24.04$&$\color{red}57.06\pm32.36$&$\color{red}57.70\pm26.98$&$\color{red}47.14\pm26.27$&$\color{red}48.25\pm31.01$&$\color{red}44.81\pm26.47$&$\color{red}38.87\pm21.98$&$\color{red}48.98\pm28.92$\\
Mistral-7B-Instruct &$\color{red}59.97\pm31.97$&$\color{red}47.52\pm27.06$&$\color{red}63.69\pm32.38$&$\color{red}67.63\pm21.04$&$\color{red}53.04\pm17.28$&$\color{red}56.98\pm24.03$&$\color{red}51.95\pm27.87$&$\color{red}58.10\pm24.20$&$\color{red}63.61\pm25.67$\\
Meta-LLaMA-3-8B-Instruct &$\color{red}78.42\pm12.33$&$\color{red}61.08\pm22.62$&$\color{red}80.36\pm15.75$&$\color{red}69.20\pm12.21$&$\color{red}66.61\pm12.08$&$\color{red}57.30\pm24.21$&$\color{red}61.36\pm21.45$&$\color{red}59.34\pm20.28$&$\color{red}67.35\pm16.40$\\
LLaMA-2-13B-chat &$\color{red}54.02\pm28.50$&$\color{red}50.80\pm25.36$&$\color{red}61.65\pm32.55$&$\color{red}69.31\pm22.56$&$\color{red}48.93\pm18.09$&$\color{red}53.33\pm29.10$&$\color{red}45.29\pm27.55$&$\color{red}44.64\pm19.47$&$\color{red}57.82\pm30.42$\\
Aya &$\color{red}52.38\pm11.52$&$\color{red}43.73\pm12.96$&$\color{red}51.11\pm16.86$&$\color{red}55.13\pm14.45$&$\color{red}36.79\pm18.11$&$\color{red}42.22\pm22.37$&$\color{red}44.97\pm19.10$&$\color{red}39.01\pm15.25$&$\color{red}42.01\pm26.08$\\
GPT-4 &$95.83\pm3.86$&$85.42\pm8.05$&$95.92\pm7.17$&$92.86\pm2.19$&$92.14\pm3.64$&$\color{red}79.68\pm21.43$&$\color{red}93.51\pm10.83$&$93.96\pm7.65$&$\color{red}93.20\pm11.05$\\
Mixtral-8x7B &$\color{red}67.86\pm25.99$&$\color{red}62.03\pm25.21$&$\color{red}70.24\pm27.99$&$\color{red}67.30\pm27.96$&$\color{red}58.75\pm24.89$&$\color{red}57.94\pm31.36$&$\color{red}62.50\pm35.23$&$\color{red}57.14\pm33.89$&$\color{red}67.69\pm30.52$\\ \hline

\multicolumn{10}{c}{\cellcolor[HTML]{EEEEEE}Leader}     
 \\ \hline
Bloomz-7b1 &$34.37\pm6.99$&$\color{red}19.35\pm13.76$&$23.47\pm3.50$&$26.59\pm2.77$&$\color{red}13.93\pm11.87$&$14.64\pm5.54$&$26.70\pm8.15$&$27.57\pm3.50$&$29.20\pm3.19$\\
LLaMA-2-7B-chat&$\color{red}46.11\pm25.30$&$\color{red}31.55\pm17.59$&$\color{red}36.22\pm20.71$&$\color{red}37.20\pm22.42$&$\color{red}25.88\pm10.29$&$\color{red}30.24\pm14.99$&$\color{red}33.16\pm11.90$&$\color{red}23.49\pm10.17$&$\color{red}21.64\pm19.71$\\
Mistral-7B-Instruct &$\color{red}44.44\pm14.23$&$30.65\pm8.89$&$\color{red}35.97\pm13.88$&$28.77\pm7.77$&$23.42\pm5.12$&$\color{red}38.33\pm10.22$&$30.10\pm6.09$&$21.15\pm1.99$&$17.65\pm6.85$\\
Meta-LLaMA-3-8B-Instruct &$\color{red}76.03\pm24.05$&$\color{red}67.56\pm20.10$&$\color{red}70.92\pm20.47$&$\color{red}70.44\pm23.54$&$\color{red}56.62\pm22.94$&$\color{red}67.26\pm24.52$&$\color{red}58.16\pm14.15$&$\color{red}46.82\pm17.78$&$\color{red}53.57\pm19.57$\\
LLaMA-2-13B-chat &$\color{red}29.05\pm15.14$&$\color{red}36.01\pm18.25$&$\color{red}47.70\pm17.38$&$\color{red}32.14\pm13.35$&$26.64\pm4.65$&$\color{red}33.57\pm11.67$&$39.80\pm8.53$&$25.00\pm7.01$&$\color{red}24.37\pm15.70$\\
Aya &$22.30\pm6.50$&$26.49\pm5.76$&$24.74\pm4.21$&$30.36\pm3.40$&$13.41\pm1.89$&$14.76\pm2.21$&$11.05\pm3.58$&$23.94\pm1.82$&$40.34\pm5.93$\\
GPT-4 &$80.95\pm4.42$&$\color{red}63.10\pm18.29$&$\color{red}56.12\pm13.99$&$52.78\pm6.30$&$45.67\pm8.52$&$59.05\pm9.87$&$\color{red}50.34\pm10.45$&$\color{red}46.21\pm10.36$&$\color{red}55.46\pm10.83$\\
Mixtral-8x7B &$\color{red}69.37\pm34.96$&$\color{red}58.63\pm26.87$&$\color{red}55.10\pm26.99$&$\color{red}64.58\pm34.36$&$\color{red}58.67\pm34.88$&$\color{red}61.43\pm33.31$&$\color{red}52.04\pm28.85$&$\color{red}50.00\pm26.65$&$\color{red}54.62\pm30.31$\\ \hline

\multicolumn{10}{c}{\cellcolor[HTML]{EEEEEE}Wars}     
 \\ \hline
Bloomz-7b1 &$39.58\pm5.78$&$\color{red}29.22\pm10.96$&$34.69\pm4.55$&$30.47\pm7.25$&$2\color{red}6.33\pm12.41$&$37.32\pm9.33$&$22.45\pm5.84$&$35.87\pm5.36$&$28.91\pm3.66$\\
LLaMA-2-7B-chat&$\color{red}41.31\pm10.99$&$58.12\pm8.83$&$44.50\pm9.50$&$\color{red}41.22\pm15.10$&$\color{red}51.66\pm10.83$&$\color{red}63.05\pm10.14$&$22.85\pm7.31$&$45.50\pm4.39$&$\color{red}35.18\pm9.75$\\
Mistral-7B-Instruct &$42.42\pm9.40$&$\color{red}56.49\pm24.37$&$37.89\pm7.93$&$\color{red}46.68\pm10.50$&$46.84\pm8.23$&$\color{red}60.84\pm16.54$&$27.10\pm9.14$&$40.53\pm6.70$&$34.99\pm7.03$\\
Meta-LLaMA-3-8B-Instruct &$61.78\pm6.96$&$\color{red}70.13\pm12.47$&$58.13\pm8.80$&$61.30\pm9.38$&$65.78\pm7.56$&$\color{red}69.83\pm10.63$&$\color{red}51.93\pm12.07$&$53.34\pm7.37$&$52.13\pm6.86$\\
LLaMA-2-13B-chat &$44.16\pm9.90$&$34.42\pm9.98$&$37.11\pm7.29$&$41.03\pm7.96$&$37.46\pm8.56$&$\color{red}47.78\pm10.86$&$25.96\pm8.24$&$34.86\pm4.88$&$33.25\pm7.40$\\
Aya &$28.81\pm2.35$&$13.31\pm4.92$&$23.81\pm3.58$&$28.43\pm3.45$&$35.38\pm5.23$&$38.79\pm4.07$&$11.17\pm2.72$&$24.53\pm4.54$&$29.26\pm1.53$\\
GPT-4 &$79.54\pm6.02$&$93.51\pm4.11$&$71.43\pm4.41$&$76.53\pm6.61$&$82.39\pm4.79$&$78.33\pm9.53$&$31.07\pm7.77$&$56.83\pm7.30$&$59.33\pm6.72$\\
Mixtral-8x7B &$\color{red}60.23\pm17.63$&$\color{red}73.05\pm24.55$&$\color{red}62.64\pm16.80$&$\color{red}63.45\pm19.24$&$\color{red}69.44\pm18.05$&$\color{red}66.01\pm16.65$&$\color{red}55.90\pm21.42$&$\color{red}50.08\pm13.36$&$\color{red}53.80\pm15.92$\\ \hline

\multicolumn{10}{c}{\cellcolor[HTML]{EEEEEE}National Parks}     
 \\ \hline
Bloomz-7b1 &$\color{red}43.15\pm22.78$&$\color{red}16.43\pm19.95$&$\color{red}29.96\pm10.43$&$\color{red}54.76\pm35.01$&$16.70\pm9.41$&$\color{red}24.76\pm10.25$&$27.56\pm9.69$&$\color{red}29.69\pm13.11$&$\color{red}48.66\pm32.61$\\
LLaMA-2-7B-chat&$\color{red}52.94\pm27.82$&$\color{red}38.57\pm23.41$&$\color{red}47.12\pm25.45$&$\color{red}53.81\pm30.65$&$\color{red}46.95\pm21.53$&$\color{red}36.19\pm17.65$&$\color{red}32.38\pm14.22$&$\color{red}31.81\pm15.33$&$\color{red}45.54\pm27.04$\\
Mistral-7B-Instruct &$\color{red}59.50\pm25.86$&$\color{red}44.29\pm29.93$&$\color{red}49.14\pm29.51$&$\color{red}79.52\pm27.58$&$\color{red}40.23\pm20.14$&$\color{red}47.62\pm25.69$&$\color{red}36.96\pm17.38$&$\color{red}32.14\pm20.54$&$\color{red}37.28\pm27.26$\\
Meta-LLaMA-3-8B-Instruct &$\color{red}66.01\pm11.78$&$\color{red}60.00\pm22.99$&$\color{red}62.09\pm20.36$&$79.29\pm8.16$&$\color{red}63.03\pm15.82$&$\color{red}36.90\pm21.67$&$\color{red}51.63\pm18.15$&$\color{red}45.54\pm19.77$&$\color{red}53.79\pm15.49$\\
LLaMA-2-13B-chat &$\color{red}53.40\pm28.04$&$\color{red}48.57\pm19.77$&$\color{red}47.92\pm27.22$&$\color{red}70.48\pm24.44$&$\color{red}56.09\pm20.98$&$\color{red}36.90\pm14.65$&$\color{red}32.61\pm15.85$&$\color{red}34.71\pm18.00$&$\color{red}38.62\pm23.32$\\
Aya &$\color{red}42.86\pm13.39$&$\color{red}24.29\pm25.13$&$40.44\pm9.90$&$\color{red}59.29\pm15.68$&$39.29\pm7.23$&$22.86\pm7.90$&$\color{red}30.36\pm10.76$&$31.92\pm8.06$&$\color{red}32.14\pm20.66$\\
GPT-4 &$93.32\pm4.98$&$\color{red}85.71\pm23.21$&$\color{red}87.30\pm15.09$&$98.10\pm3.01$&$87.82\pm5.08$&$\color{red}66.67\pm33.24$&$\color{red}73.29\pm18.62$&$\color{red}76.56\pm19.30$&$\color{red}83.04\pm17.90$\\
Mixtral-8x7B &$\color{red}63.42\pm24.37$&$\color{red}52.86\pm28.27$&$\color{red}60.48\pm32.37$&$\color{red}72.38\pm23.67$&$\color{red}60.71\pm28.82$&$\color{red}49.05\pm23.48$&$\color{red}50.00\pm29.42$&$\color{red}49.00\pm29.10$&$\color{red}58.93\pm34.04$\\ \hline
\end{tabular}
}\vspace{-2ex}
\caption{Accuracy and standard deviation of accuracy across the seven languages covered by \datasetname\ on the countries covered. All models show high variation in accuracy over languages. Results with standard deviation higher than ten are marked with \textcolor{red}{red} color.}
\label{tab:sd}
\vspace{-5mm}
\end{table*}

\begin{figure*}[!t]
\begin{center}
\includegraphics[width=\textwidth]{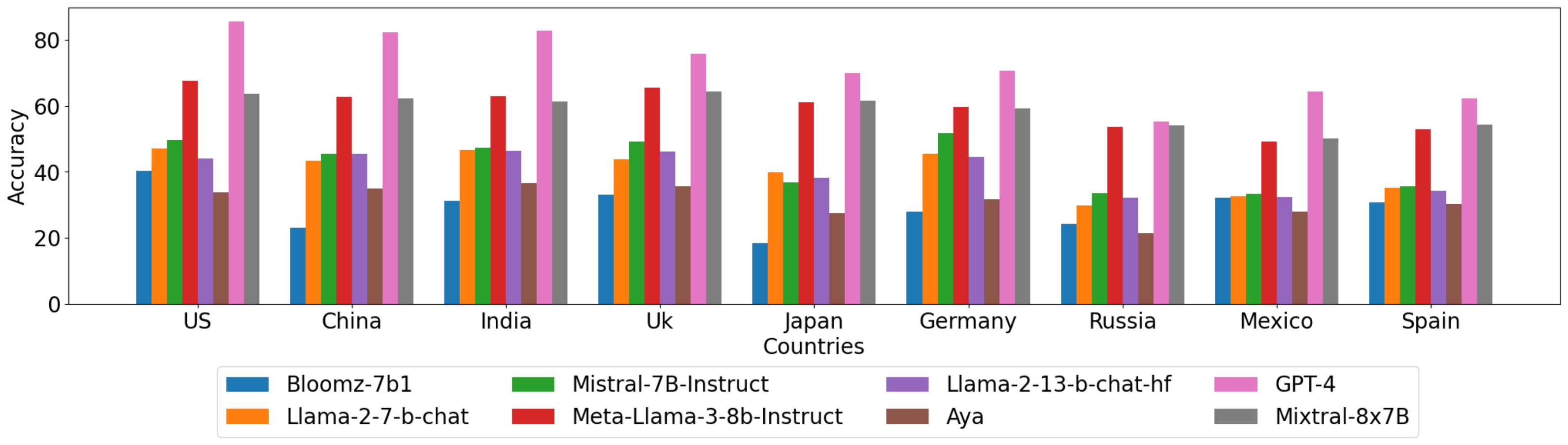}
\caption{Accuracy of models across nations averaged over language. We note that models like GPT-4 demonstrate relatively low performance for Mexico, Spain, and Russia as compared to India and China even though the native languages of these nations (Hindi and Chinese) are poorly performing (c.f. Table~\ref{tab:main_results}). 
}
\label{fig:country_performance}
\end{center}
\vspace{-5mm}
\end{figure*}

\section{Evaluation of Qwen3 and LLaMA-3.1 Models} 
\label{sec:eval_qwen_llama3.1_appendix}
Table~\ref{tab:new_results} presents the performance of Meta-Llama-3.1-8B-Instruct~\citep{meta2024llama3.1}, Qwen3-8B~\citep{qwen3}, and Qwen3-14B~\citep{qwen3}. Similar to other models, we observe that they achieve stronger results on Western languages. Moreover, while Meta-Llama-3.1-8B-Instruct performs best among them, its performance still lags behind GPT-4.

\begin{table*}[!t]
\centering
\resizebox{0.9\textwidth}{!}{
\begin{tabular}{lccccccc|ccc}
\hline
 \backslashbox{Model}{Lang}& EN& DE& ES& HI& RU& JA &ZH & AVG& $\text{AVG}_{W}$&$\text{AVG}_{NW}$\\ 
 \hline
\multicolumn{11}{c}{\cellcolor[HTML]{EEEEEE}Monuments} \\ \hline
Meta-LLaMA-3.1-8B-Instruct & 37.01 & \textbf{73.31} & \textbf{78.47} & \textbf{51.25} & \textbf{68.95} & \textbf{53.74} & \textbf{58.81} & \textbf{60.22} & \textbf{64.44} & \textbf{54.59} \\
Qwen3-8B                  & 21.44 & 59.96 & 55.87 & 17.70 & 35.05 & 41.99 & 26.96 & 37.00 & 43.08 & 28.89 \\
Qwen3-14B                 & \textbf{57.92} & 61.21 & 77.14 & 53.65 & 46.17 & 29.89 & 42.62 & 52.66 & 60.61 & 42.06 \\
\hline
\multicolumn{11}{c}{\cellcolor[HTML]{EEEEEE}Leaders} \\ \hline
Meta-LLaMA-3.1-8B-Instruct & \textbf{72.75} & \textbf{76.50} & \textbf{76.58} & \textbf{49.75} & \textbf{58.83} & \textbf{50.00} & \textbf{31.50} & \textbf{59.42} & \textbf{71.17} & \textbf{43.75} \\
Qwen3-8B                  & 12.00 & 40.17 & 48.33 & 30.25 & 43.25 & 15.33 & 14.08 & 29.06 & 35.94 & 19.88 \\
Qwen3-14B                 & 63.75 & 59.58 & 59.08 & 31.67 & 50.25 & 3.75 & 11.67 & 39.96 & 58.17 & 15.68 \\
\hline
\multicolumn{11}{c}{\cellcolor[HTML]{EEEEEE}Wars} \\ \hline
Meta-LLaMA-3.1-8B-Instruct & 66.47 & 64.84 & 62.68 & \textbf{55.63} & 60.80 & \textbf{58.07} & \textbf{54.04} & \textbf{60.36} & \textbf{63.70} & \textbf{55.90} \\
Qwen3-8B                  & 40.93 & 54.75 & 57.29 & 46.88 & 55.45 & 47.27 & 41.25 & 49.12 & 52.11 & 45.13 \\
Qwen3-14B                 & \textbf{67.85} & \textbf{63.42} & \textbf{63.74} & 48.97 & \textbf{59.74} & 48.09 & 41.57 & 56.20 & 63.69 & 46.21 \\
\hline
\multicolumn{11}{c}{\cellcolor[HTML]{EEEEEE}National Parks} \\ \hline
Meta-LLaMA-3.1-8B-Instruct & 39.48 & 57.45 & 71.19 & \textbf{46.62} & \textbf{58.77} & \textbf{41.70} & \textbf{40.38} & \textbf{50.80} & \textbf{56.73} & \textbf{42.89} \\
Qwen3-8B                  & 26.69 & 57.51 & 54.12 & 20.56 & 26.69 & 27.96 & 19.56 & 33.30 & 41.25 & 22.70 \\
Qwen3-14B                 & \textbf{72.04} & \textbf{57.03} & \textbf{77.54} & 44.82 & 50.26 & 21.72 & 30.50 & 50.56 & 64.22 & 32.34 \\
\hline
\end{tabular}
}
\caption{Accuracy of Meta-LLaMA-3.1 and Qwen3 models across languages and domains. Columns $\text{AVG}_{W}$ and $\text{AVG}_{NW}$ show average performance over Western (EN, DE, ES, RU) and non-Western (HI, JA, ZH) languages respectively. Meta-LLaMA-3.1-8B-Instruct consistently outperforms Qwen3 models across most domains and languages, except in some cases where Qwen3-14B shows higher scores in EN and ES.}
\label{tab:new_results}
\vspace{-5mm}
\end{table*}

\end{document}